\documentclass{article}

\usepackage{arxiv}

\usepackage[utf8]{inputenc} %
\usepackage[T1]{fontenc}    %
\usepackage{hyperref}       %
\usepackage{url}            %
\usepackage{booktabs}       %
\usepackage{amsfonts}       %
\usepackage{nicefrac}       %
\usepackage{microtype}      %
\usepackage{lipsum}         %

\newcommand{\dquote}[1]{``#1''}

\usepackage{url}            %
\usepackage{booktabs}       %
\usepackage{multirow}    
\usepackage{amsfonts}       %
\usepackage{nicefrac}       %
\usepackage{microtype}      %
\usepackage{natbib}
\usepackage{enumerate}
\usepackage{hhline}
\usepackage{makecell}
\usepackage{pifont}

\usepackage{graphicx} %
\usepackage{caption}
\usepackage{subcaption}
\usepackage{amsmath}
\usepackage{amsthm}
\usepackage{amssymb}
\usepackage{tikz}
\usepackage{xcolor}
\usetikzlibrary{arrows}

\allowdisplaybreaks

\usepackage{mathrsfs}

\usepackage{algorithm}
\usepackage{algorithmic}
\usepackage{hyperref}
\usepackage{bm}

\allowdisplaybreaks

\newcommand{\labs}{\left\vert}
\newcommand{\rabs}{\right\vert}
\newcommand{\lnorm}{\left\Vert}
\newcommand{\rnorm}{\right\Vert}

\newcommand{\expect}{\mathbb{E}}

\newtheorem{thm}{Theorem}

\newtheorem{defn}{Definition}

\usepackage[capitalize,noabbrev]{cleveref}
\crefname{thm}{Theorem}{Theorems}
\crefname{lem}{Lemma}{Lemmas}
\crefname{cor}{Corollary}{Corollaries}
\crefname{prop}{Proposition}{Propositions}
\crefname{asmp}{Assumption}{Assumptions}
\crefname{defn}{Definition}{Definitions}
\crefname{oracle}{Oracle}{Oracles}
\crefname{fact}{Fact}{Facts}
\crefname{conj}{Conjecture}{Conjectures}
\crefname{rem}{Remark}{Remarks}
\crefname{claim}{Claim}{Claims}

\definecolor{red}{rgb}{1, 0, 0}

\definecolor{green}{rgb}{0, 1, 0}

\definecolor{blue}{rgb}{0, 0, 1}

\definecolor{orange}{rgb}{1, 0.4, 0.0}

\usepackage{amsmath,amsfonts,bm}
\makeatletter
\let\save@mathaccent\mathaccent
\newcommand*\if@single[3]{%
  \setbox0\hbox{${\mathaccent"0362{#1}}^H$}%
  \setbox2\hbox{${\mathaccent"0362{\kern0pt#1}}^H$}%
  \ifdim\ht0=\ht2 #3\else #2\fi
  }
\newcommand*\rel@kern[1]{\kern#1\dimexpr\macc@kerna}
\newcommand*\widebar[1]{\@ifnextchar^{{\wide@bar{#1}{0}}}{\wide@bar{#1}{1}}}
\newcommand*\wide@bar[2]{\if@single{#1}{\wide@bar@{#1}{#2}{1}}{\wide@bar@{#1}{#2}{2}}}
\newcommand*\wide@bar@[3]{%
  \begingroup
  \def\mathaccent##1##2{%
    \let\mathaccent\save@mathaccent
    \if#32 \let\macc@nucleus\first@char \fi
    \setbox\z@\hbox{$\macc@style{\macc@nucleus}_{}$}%
    \setbox\tw@\hbox{$\macc@style{\macc@nucleus}{}_{}$}%
    \dimen@\wd\tw@
    \advance\dimen@-\wd\z@
    \divide\dimen@ 3
    \@tempdima\wd\tw@
    \advance\@tempdima-\scriptspace
    \divide\@tempdima 10
    \advance\dimen@-\@tempdima
    \ifdim\dimen@>\z@ \dimen@0pt\fi
    \rel@kern{0.6}\kern-\dimen@
    \if#31
      \overline{\rel@kern{-0.6}\kern\dimen@\macc@nucleus\rel@kern{0.4}\kern\dimen@}%
      \advance\dimen@0.4\dimexpr\macc@kerna
      \let\final@kern#2%
      \ifdim\dimen@<\z@ \let\final@kern1\fi
      \if\final@kern1 \kern-\dimen@\fi
    \else
      \overline{\rel@kern{-0.6}\kern\dimen@#1}%
    \fi
  }%
  \macc@depth\@ne
  \let\math@bgroup\@empty \let\math@egroup\macc@set@skewchar
  \mathsurround\z@ \frozen@everymath{\mathgroup\macc@group\relax}%
  \macc@set@skewchar\relax
  \let\mathaccentV\macc@nested@a
  \if#31
    \macc@nested@a\relax111{#1}%
  \else
    \def\gobble@till@marker##1\endmarker{}%
    \futurelet\first@char\gobble@till@marker#1\endmarker
    \ifcat\noexpand\first@char A\else
      \def\first@char{}%
    \fi
    \macc@nested@a\relax111{\first@char}%
  \fi
  \endgroup
}
\makeatother

\def\eqref#1{(\ref{#1})}

\def\1{\bm{1}}

\DeclareMathAlphabet{\mathsfit}{\encodingdefault}{\sfdefault}{m}{sl}
\SetMathAlphabet{\mathsfit}{bold}{\encodingdefault}{\sfdefault}{bx}{n}

\def\gA{{\mathcal{A}}}

\def\gN{{\mathcal{N}}}

\def\gS{{\mathcal{S}}}

\def\gX{{\mathcal{X}}}

\def\sP{{\mathbb{P}}}

\newcommand{\e}{\begin{equation}}
\newcommand{\ee}{\end{equation}}
\newcommand{\ea}{\begin{equation*}}
\newcommand{\eea}{\end{equation*}}

\newcommand{\lp}{\left(}
\newcommand{\rp}{\right)}

\newcommand{\ls}{\left[}
\newcommand{\rs}{\right]}

\usepackage{todonotes}

\title{A Survey on Model-based Reinforcement Learning}

\AtBeginDocument{\setlength\abovedisplayskip{4pt}}
\AtBeginDocument{\setlength\belowdisplayskip{4pt}}

\renewcommand{\textsf}[1]{#1}

\usepackage{framed}
\colorlet{shadecolor}{orange!15}

\usepackage{enumitem}
\setlist{topsep=0pt,parsep=0pt,partopsep=0pt}

\usepackage{tabulary}

\hypersetup{
    colorlinks=true,
    citecolor=blue,
    linkcolor=blue,
}

\theoremstyle{plain}

\theoremstyle{definition}

\theoremstyle{remark}

\usepackage{authblk}
\newcommand\nnfootnote[1]{%
  \begin{NoHyper}
  \renewcommand\thefootnote{}\footnote{#1}%
  \addtocounter{footnote}{-1}%
  \end{NoHyper}
}
\author[1,3]{\textbf{Fan-Ming Luo}}
\author[1]{\textbf{Tian Xu}}
\author[2]{\textbf{Hang Lai}}
\author[1,3]{\textbf{Xiong-Hui Chen}}
\author[2]{\textbf{Weinan Zhang}\textsuperscript{\dag}}
\author[1,3]{\textbf{Yang Yu}\textsuperscript{\dag}}
\affil[1]{National Key Laboratory for Novel Software Technology, Nanjing University, China}
\affil[2]{Shanghai Jiao Tong University, China}
\affil[3]{Polixir.ai}
\affil[ ]{\texttt{ \{luofm,xut,chenxh,yuy\}@lamda.nju.edu.cn,laihang@apex.sjtu.edu.cn,wnzhang@sjtu.edu.cn}}

\renewcommand{\headeright}{}
\renewcommand{\shorttitle}{}

\date{}
\begin{document}
\maketitle

\begin{abstract}
Reinforcement learning (RL) solves sequential decision-making problems via a trial-and-error  process interacting with the environment. While RL achieves outstanding success in playing complex video games that allow huge trial-and-error,  making errors is always undesired in the real world. To improve the sample efficiency and thus reduce the errors, model-based reinforcement learning (MBRL) is believed to be a promising direction, which builds environment models in which the trial-and-errors can take place without real costs. In this survey, we take a review of MBRL with a focus on the recent progress in deep RL.
For non-tabular environments, there is always a generalization error between the learned environment model and the real environment. As such, it is of great importance to analyze the discrepancy between policy training in the environment model and that in the real environment, which in turn guides the algorithm design for better model learning, model usage, and policy training. Besides, we also discuss the recent advances of model-based techniques in other forms of RL, including offline RL, goal-conditioned RL, multi-agent RL, and meta-RL. Moreover, we discuss the applicability and advantages of MBRL in real-world tasks. Finally, we end this survey by discussing the promising prospects for the future development of MBRL. We think that MBRL has great potential and advantages in real-world applications that were overlooked, and we hope this survey could attract more research on MBRL.
\end{abstract}

\nnfootnote{\dag: Corresponding authors.}


\section{Overview of Model-based RL}

Reinforcement learning (RL) studies the methodologies of improving the performance of sequential decision-making for autonomous agents \citep{sutton2018reinforcement}. Since the success of deep RL in playing the game of Go and video games shows the beyond-human ability of decision-making, it is of great interest to extend its application horizon to include real-world tasks.

Typically, deep RL algorithms require tremendous training samples, resulting in much high sample complexity. In general RL tasks, the sample complexity of a particular algorithm refers to the amount of samples required for learning an approximately optimal policy. Particularly, unlike the supervised learning paradigm that learns from historical labeled data, typical RL algorithms require the interaction data by running the latest policy in the environment. Once the policy updates, the underlying data distribution (formally the occupancy measure \citep{syed2008apprenticeship}) changes, and the data has to be collected again by running the policy. As such, RL algorithms with high sample complexity are hard to be directly applied in real-world tasks, where trial-and-errors can be highly costly.

Therefore, a major focus of the recent research on deep reinforcement learning (DRL) is on improving sample efficiency \citep{yu2018towards}. Among different branches of research, model-based reinforcement learning (MBRL) is one of the most important directions that is widely believed to have the great potential to make RL algorithms significantly more sample efficient \citep{wang2019benchmarking}. This belief is intuitively from an analogy with human intelligence. Human beings are capable of having an imagined world in mind, in which how things could happen following different actions can be predicted. In such a way, proper actions can be chosen to take from imagination and are thus with low trial-and-error costs. The phrase \emph{model} in MBRL is the environment model that is expected to play the same role as the imagination.


In MBRL, the environment model (or simply the model) refers to the abstraction of the environment dynamics with which the learning agent interacts. The dynamics environment in RL is typically formulated as a Markov decision process (MDP), denoted with a tuple $\langle S, A, M, R, \gamma \rangle$, where $S$, $A$ and $\gamma$ denote the state space, action space and the discount factor for future rewards, respectively, while $M: S \times A \mapsto S$ denotes the state transition dynamics and $R: S \times A \mapsto \mathbb{R}$ denotes the reward function. Normally, given the state and action spaces and the discount factor, the key components of the environment model are the state transition dynamics and the reward function. Thus, learning the model corresponds to recovering the state transition dynamics $M$ and the reward function $R$. In many cases, the reward function is also explicitly defined, thus the major task of the model learning is to learn the state transition dynamics \citep{slbo,mbpo}.

\begin{figure}[b]
    \centering
    \begin{minipage}[c]{0.32\linewidth}
    \centering
    \includegraphics[width=\linewidth]{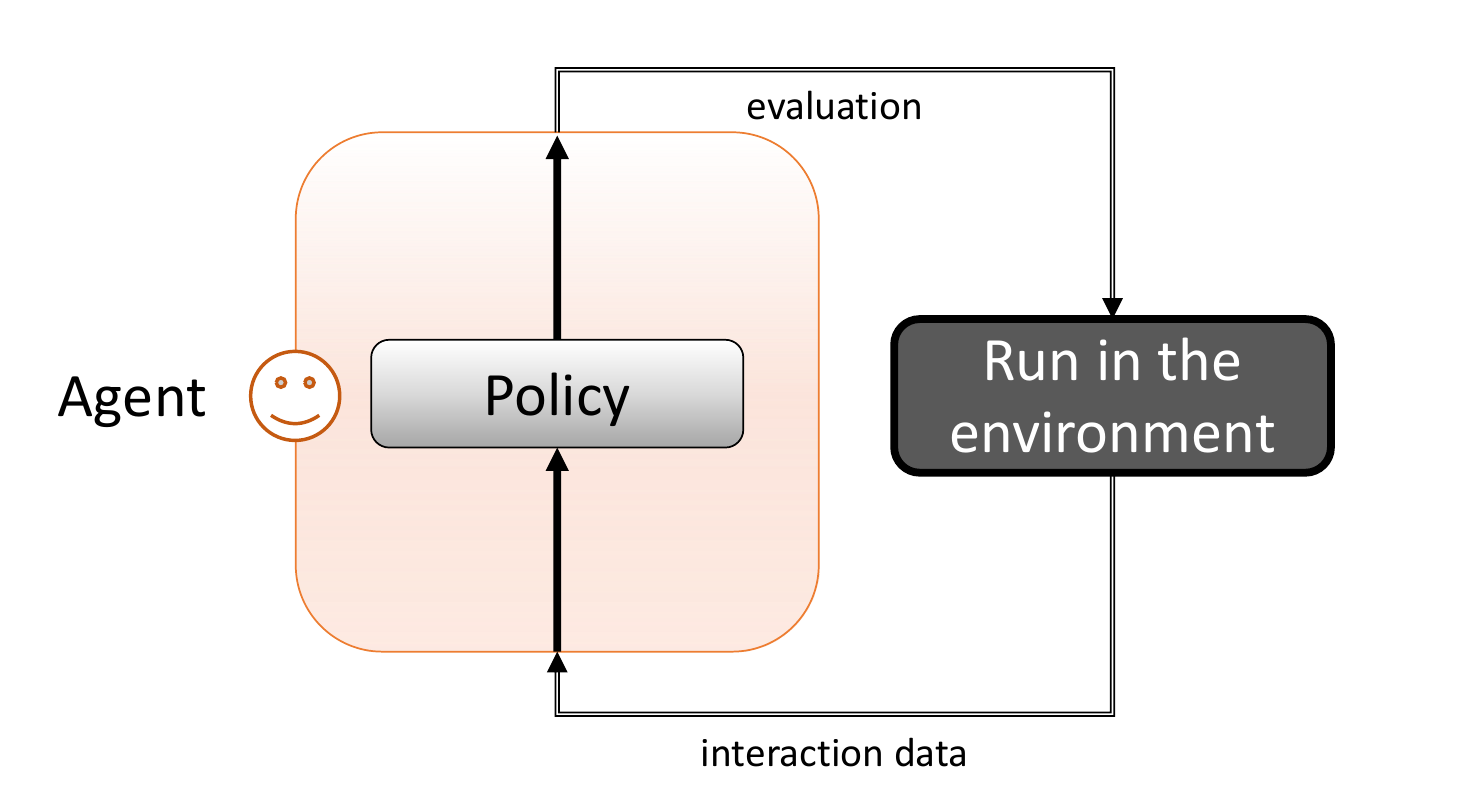}\\
    (a) on-policy RL
    \end{minipage}
    \begin{minipage}[c]{0.32\linewidth}
    \centering
    \includegraphics[width=\linewidth]{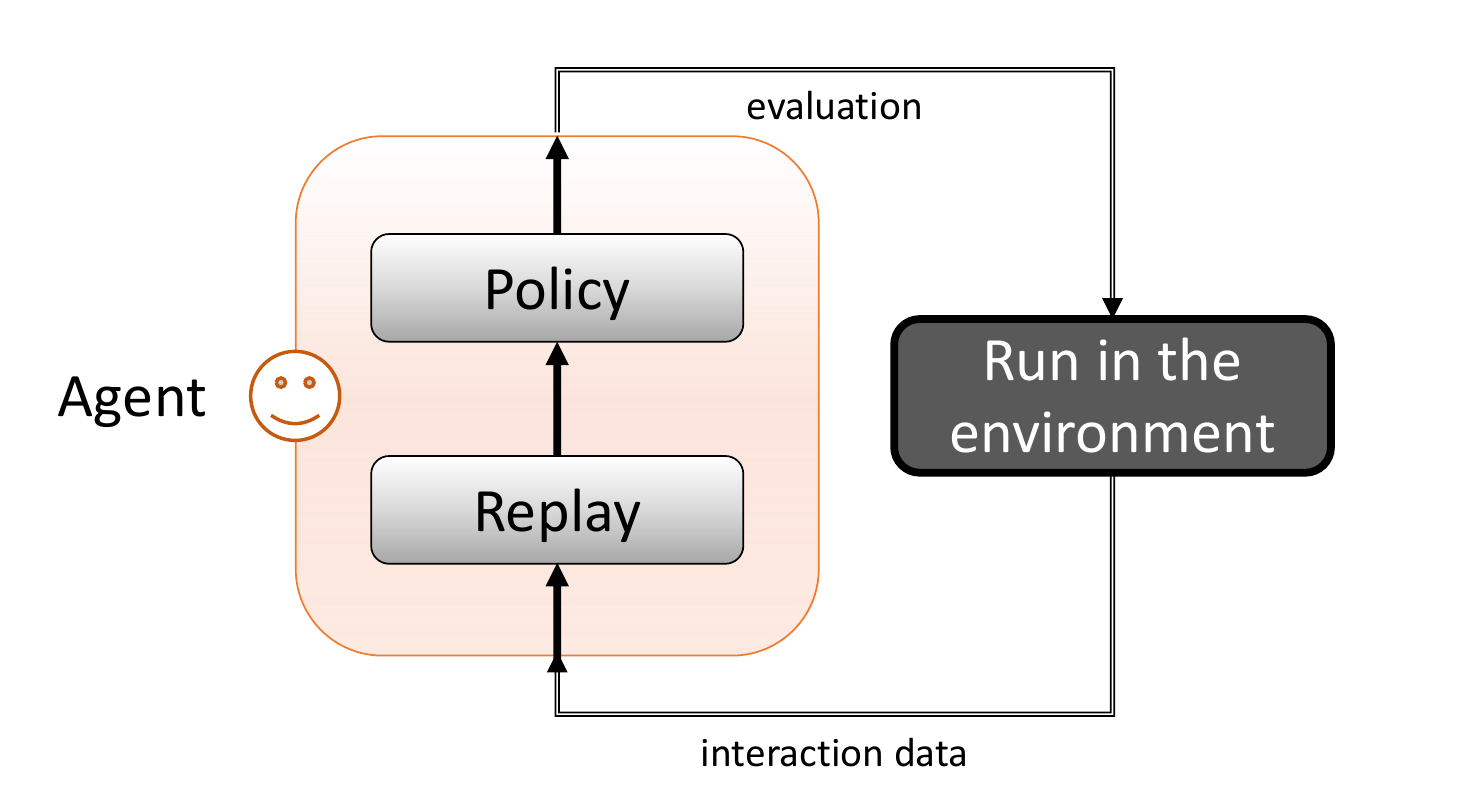}\\
    (b) off-policy RL
    \end{minipage}
    \begin{minipage}[c]{0.32\linewidth}
    \centering
    \includegraphics[width=\linewidth]{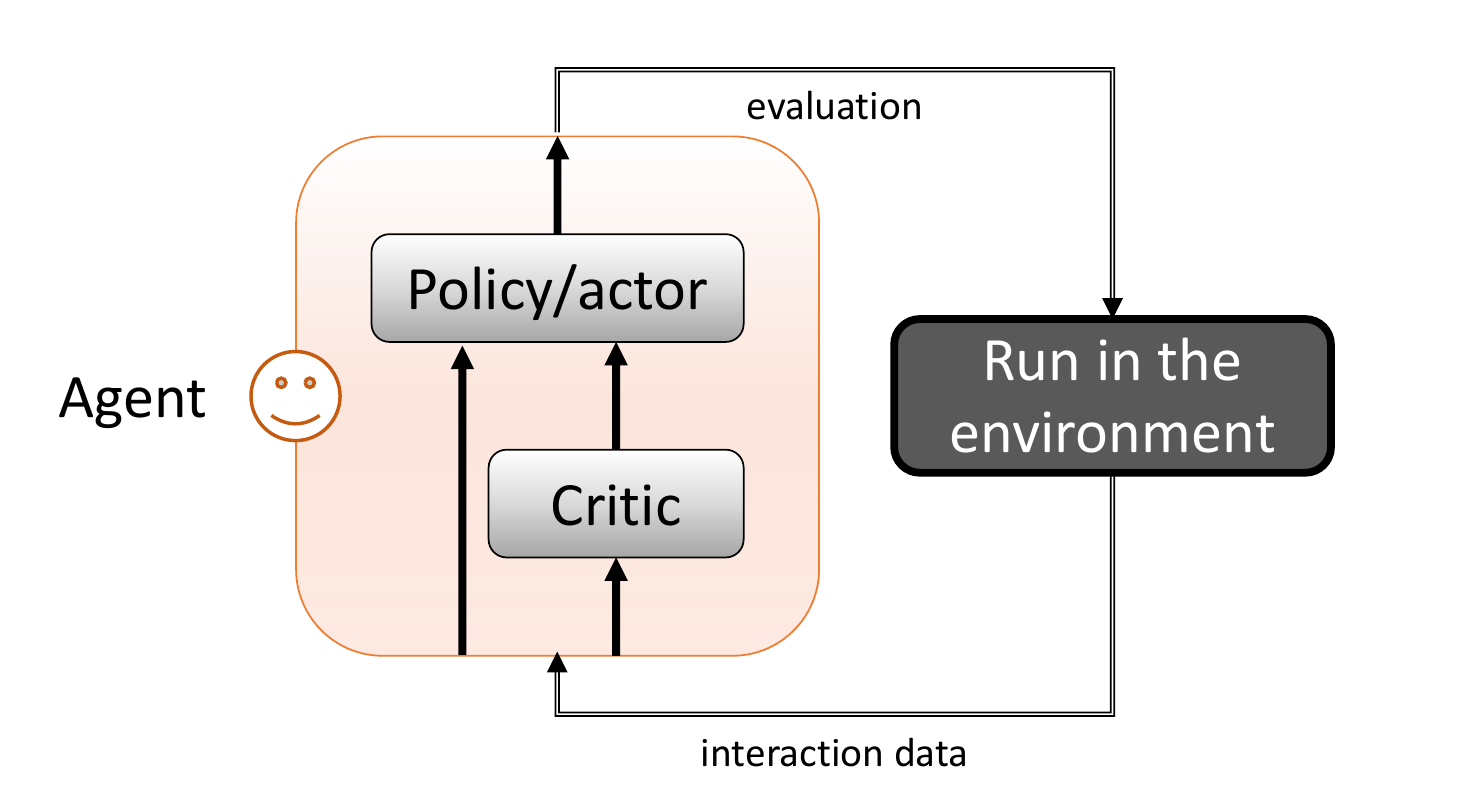}\\
    (c) actor-critic RL
    \end{minipage}\\
    \begin{minipage}[c]{0.32\linewidth}
    \centering
    \includegraphics[width=\linewidth]{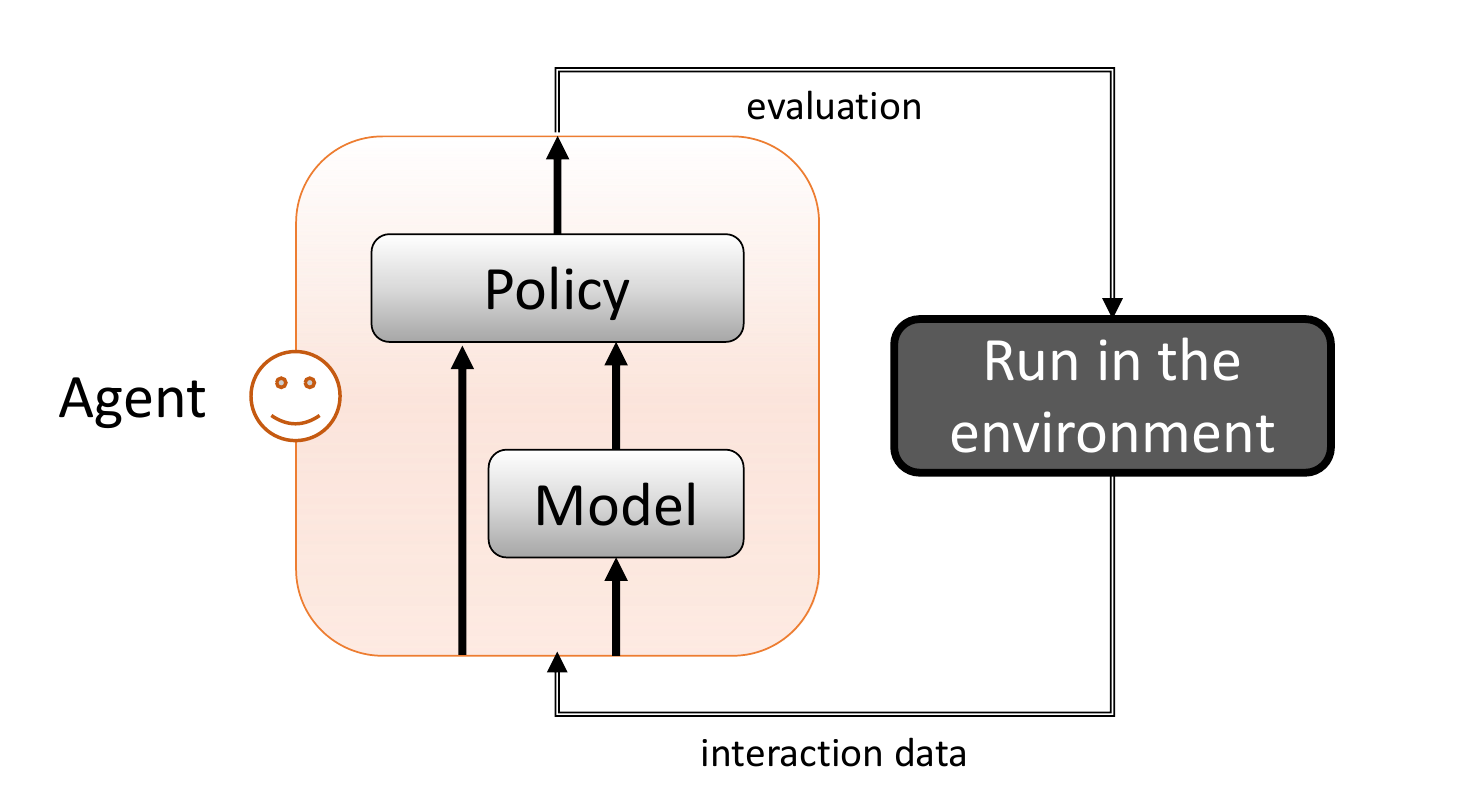}\\
    (d) model-based RL
    \end{minipage}
    \begin{minipage}[c]{0.32\linewidth}
    \centering
    \includegraphics[width=\linewidth]{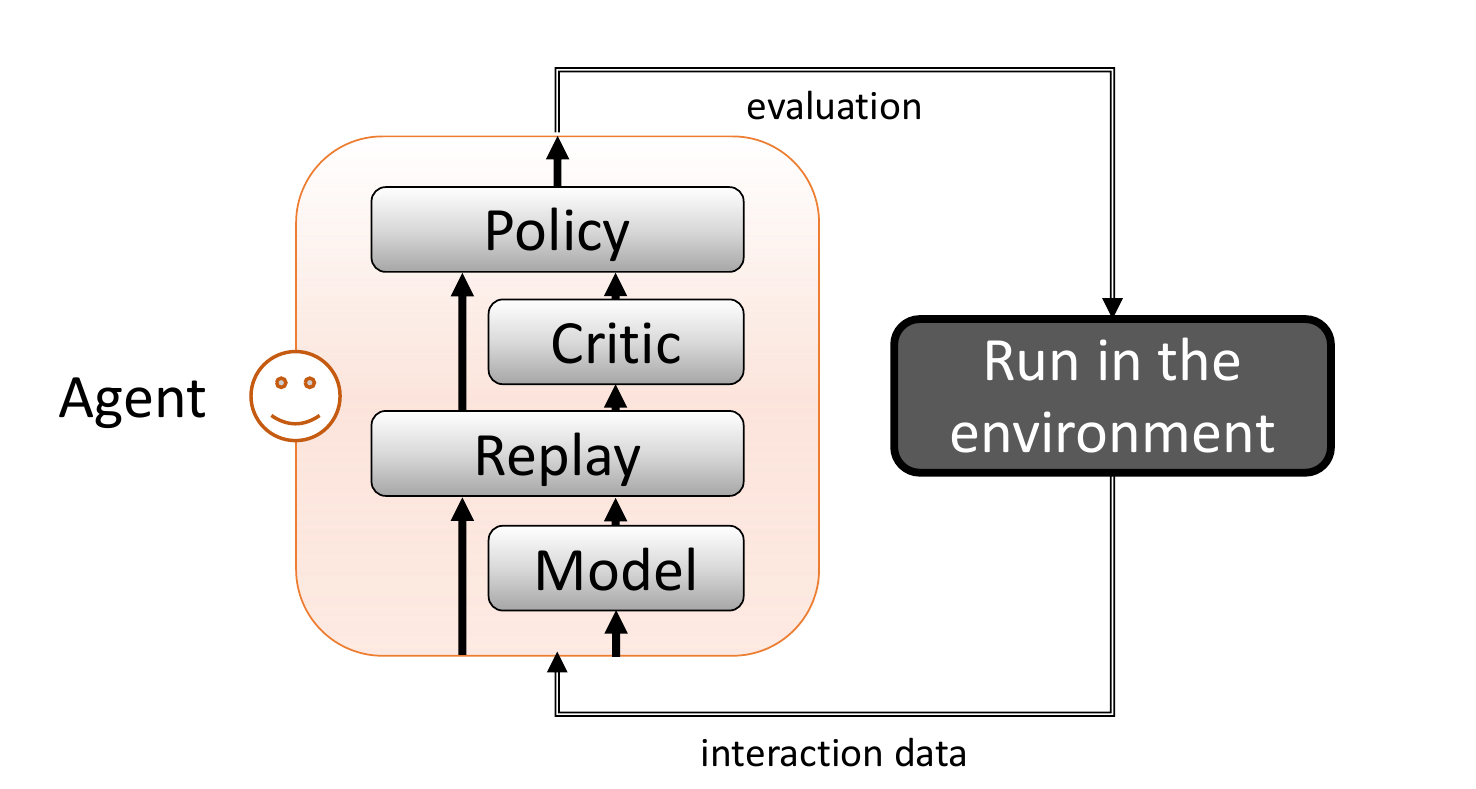}\\
    (e) off-policy actor-critic RL
    \end{minipage}
    \caption{Architectures of RL algorithms. The figures show a training iteration of the RL, focusing on how the interaction data is utilized.}
    \label{fig:RLtypes}
\end{figure}

With an environment model, the agent can have the imagination ability. It can interact with the model in order to sample the interaction data, which is also called \emph{simulation data}. Ideally, if the model is sufficiently accurate, a good policy can be learned in the model. Compared with the model-free reinforcement learning (MFRL) methods, where the agent can only use the data sampled from the interaction with the real environment, called \emph{experienced data}, MBRL enables the agent to fully leverage the experienced data in the learned model. It should be noticed that, besides MBRL, there are other approaches trying to better utilize the experienced data, such as the off-policy algorithms that employ a replay buffer to record the old data and the actor-critic algorithms that can be viewed as learning a critic to facilitate the policy updates. Figure \ref{fig:RLtypes} depicts different types of RL structures. Figure \ref{fig:RLtypes}(a) is the simplest on-policy RL, where the agent uses the latest data to update the policy. In the off-policy, as Figure \ref{fig:RLtypes}(b), the agent collects historical data in the replay buffer, in which the policy is learned. In actor-critic RL, as shown in \ref{fig:RLtypes}(c) the agent learns a critic, which is the value function of the long-term return, and then learns the policy (actor) assisted by the critic. MBRL, as shown in Figure \ref{fig:RLtypes}(d), explicitly learns a model. Compared with off-policy RL, MBRL reconstructs the state transition dynamics, while off-policy RL simply uses the replay buffer to estimate the value more robustly. Although calculating the value function, or the critic, involves the information of the transition dynamics, the learned model in MBRL is decoupled with the policy and thus can be used to evaluate other policies, while the value function is bound to the sampling policy. Also, note that off-policy, actor-critic, and model-based are three structures in parallel, Figure \ref{fig:RLtypes}(e) shows a possible combination of them. 

With a sufficiently accurate model, it is intuitive that MBRL yields higher sample efficiency than MFRL, as shown in recent studies from both theoretical \citep{sun2019model} and empirical \citep{mbpo,wang2019benchmarking} perspectives.
However, in a wide range of DRL tasks with relatively complex environments, it is not straightforward to learn an ideal model. Therefore, we need to carefully consider approaches to model learning and model usage. 

In this survey, we take a comprehensive review of the model-based reinforcement learning methods. Firstly, we focus on how models are learned and used in the basic setting, as in Sec.3 for model learning and Sec.4 for model usage.  
For \emph{model learning}, we start from the classic tabular represented models, then for approximation models such as using neural networks, we review the theories and the key challenges when facing complex environments, as well as advances for reducing model errors.
For \emph{model usage}, we categorize the literature into two parts, i.e., the blackbox model rollout for trajectory sampling and whitebox model for gradient propagation. 
Regarding model usage as a subsequent task of model learning, we also cover the attempts to bridge model learning and model usage, i.e., value-aware and policy-aware model learning.
Moreover, we take a brief review on the combination of model-based methods in other forms of reinforcement learning, including offline RL, goal-conditioned RL, multi-agent RL, and meta RL. We also discuss the applicability and advantages of MBRL in real-world tasks. 
We finally conclude this paper with a discussion of promising the emerging research perspectives and future trends in the development of MBRL.








\section{Model Learning}\label{sec:model-learning}

For MBRL, the first component to consider is the learning of the environment model. As introduced above, the model is formulated as the MDP  $\langle S, A, M, R, \gamma \rangle$, where $S$, $A$, and $\gamma$ are commonly predefined, and the state transition dynamics $M$ and the reward function $R$ are to be learned. We assume that some historical data is available for learning. Usually, the historical data can be in the form of trajectories $\{\tau_1, \tau_2, \ldots, \tau_k\}$, and each trajectory is a sequence of state-action-reward pairs, $\tau_i = (s^i_0, a^i_0, r^i_0, \ldots, s^i_{L-1}, a^i_{L-1}, r^i_{L-1}, s^i_L)$. It is easy to discover that the data records the past transitions $(s_t, a_t, s_{t+1})$ corresponding to the input and output of the transition dynamics $M$, and the past rewards $(s_t, a_t, r_t)$ corresponding to the input and output of the reward function $R$. Therefore, it is straightforward to borrow ideas from supervised learning for model learning.             

\subsection{Model Learning in Tabular Setting}

At the early stage of RL research, the state and action spaces are finite and small, the model learning is considered with the tabular MDPs \citep{Sutton90QPlanning}, where policies, transition dynamics, and reward functions can all be recorded in a table. To learn the transition dynamics, let $C[s,a,s']$ record the counting of the state-action-next-state $(s, a, s^\prime)$. The transition dynamics are then
\begin{equation}
	\hat M(s'|s,a) = \begin{cases}
		\frac{C[s,a,s']}{\sum_{s''\in S} C[s,a,s'']}, & \sum_{s''\in S} C[s,a,s'']>0\\
		\frac{1}{|S|},& \text{otherwise}
 	\end{cases}\label{eq:tabular_model}
\end{equation}
For the reward function, let $Sum[s,a]$ record the sum of rewards received by taking action $a$ on state $s$. The reward function is then
\begin{equation}
	\hat R(s,a) = \begin{cases}
		\frac{Sum[s,a]}{C[s,a]}, & C[s,a]>0\\
		R_{\min},& \text{otherwise}
 	\end{cases}
\end{equation}
where $R_{\min}$ is the preset minimum value of the immediate reward.

The above simple calculation of the transition dynamics and the reward function corresponds to the maximum likelihood estimation (MLE) under the tabular setting. Notice that $\hat{M}$ and $\hat R$ are an unbiased estimation of the true transition $M^*$ and the true reward function $R^*$ respectively, and thus converges to $M^*$ and $R^*$ as the samples approach infinity.

For collecting samples, sampling trajectories from the environment is not as straightforward as sampling a coin. R-MAX \citep{brafman02rmax} is a representative algorithm for joint model learning and exploration. In R-MAX, a state transits to itself, and the immediate reward is set to the maximum value by default, but only when a state-action pair has been visited sufficiently many times, i.e., larger than $K$, the transition probability, and the reward are assigned to their average value. This is implemented by using the $\tilde M$ and $\tilde R$ as in Eq.\eqref{eq:RMAX}.
\begin{equation}
	\tilde M(s'|s,a) = \begin{cases}
		\frac{C[s,a,s']}{\sum_{s''\in S} C[s,a,s'']}, & C[s,a]\geq K\\
		I[s'=s],& C[s,a]<K
 	\end{cases},\qquad 
	\tilde R(s,a) = \begin{cases}
		\frac{Sum[s,a]}{C[s,a]}, & C[s,a]\geq K\\
		R_{\max},& C[s,a]<K
 	\end{cases}\label{eq:RMAX}
\end{equation}
where $I$ is the indicator function that is 1 when the inner expression is true and 0 otherwise. 

In every iteration, R-MAX solves an $\epsilon$-optimal policy in the fictitious model $\langle S, A, \tilde M, \tilde R, \gamma \rangle$, which naturally tries to explore unvisited states due to the set of the $R_{\max}$ reward, and applies the policy in the environment to collect more samples to update the fictitious model. With a properly set $K$, R-MAX requires $\tilde{O}\left(\frac{|\mathcal{S}|^{2}|\mathcal{A}|}{\epsilon^{3}(1-\gamma)^{2}} \log \frac{1}{\delta}\right)$ episodes to achieve a high accuracy ($\ell_1$ difference on transition $<\epsilon/2$) model \citep{rmaxnote}.

\subsection{Model Learning via Prediction Loss}
\label{subsection:model_learning_loss_architecture_property}
While the counting method and the theory of model learning under the setting of tabular MDPs are clear, it is not feasible to use tabular representation for large-scale MDPs and MDPs with continuous state space and action space. Approximation functions are therefore employed in the general setting. The approximation functions can be implemented by machine learning models, such as linear models, neural networks, decision tree models, etc. In this survey, we focus on neural network models, for which we denote the transition model as $M_\theta$ with parameter $\theta$ being the network weights. Meanwhile, in order to explicitly indicate the real transition dynamics, we denote it as $M^*$.

\subsubsection{Prediction Model Loss} 

A straightforward approach to model learning fits one-step transitions, which has been widely employed \citep{Kurutach2018ModelEnsembleTRPO,Feinberg2018MVE, slbo, mbpo, Rajeswaran2020GameTheoretic}. When $M_\theta$ is deterministic, the model learning objective can be the mean squared prediction error of the model $M_\theta$ on the next state \citep{Nagabandi2018NNDynamicsForMBRL}.

\begin{align}
    \min_{\theta} \mathbb{E}_{(s, a) \sim \rho^{M^*}_{\pi_{D}}, s^\prime \sim M^* (\cdot|s, a)} \ls \lnorm s^\prime - M_\theta (s, a) \rnorm_{2}^2 \rs.
    \label{eq:model_objective_mse}
\end{align}
Here $M^*$ is the real transition dynamics, $\pi_{D}$ is the data-collecting policy and $\rho^{M^*}_{\pi_{D}}$ is the stationary state-action distribution induced by $\pi_{D}$ and $M^*$. Intuitively, $\rho^{M^*}_{\pi_{D}}$ is the data collected by running $\pi_D$ for a long period. 

However, the deterministic transition model fails to capture the aleatoric uncertainty \citep{Chua2018PETS}, which arises from the inherent stochasticities of the environment. To model the aleatoric uncertainty, a natural idea is to utilize the probabilistic transition model $M_\theta (\cdot|s, a)$ \citep{Chua2018PETS}. Under this case, the objective can be minimizing the KL divergence between $M^* (\cdot|s, a)$ and $M_\theta (\cdot|s, a)$ as in Eq.\eqref{eq:model_objective_kl}.    
\begin{align}
\label{eq:model_objective_kl}
    \min_{\theta} \mathbb{E}_{(s, a) \sim \rho_{\pi_{D}}^{M^*}} \left[D_{\mathrm{KL}}\left(M^*(\cdot | s, a), M_{\theta}(\cdot | s, a)\right)\right] := \mathbb{E}_{(s, a) \sim \rho_{\pi_{D}}^{M^*}, s^\prime \sim M^* (\cdot|s, a)} \ls \log \lp \frac{M^*\left(s^{\prime} \mid s, a\right)}{M_{\theta}\left(s^{\prime} \mid s, a\right)} \rp \rs .
\end{align}

The probabilistic transition model is often instantiated as a Gaussian distribution \citep{Chua2018PETS, slbo, mbpo}, i.e., $M_\theta (\cdot|s, a) = \gN (\mu_{\theta} (s, a), \Sigma_{\theta} (s, a))$ with parameterized models of $\mu_{\theta}$ and $\Sigma_{\theta}$. Then Eq.\eqref{eq:model_objective_kl} becomes
\begin{align*}
    \min_{\theta} \mathbb{E}_{(s, a) \sim \rho_{\pi_{D}}^{M^*}, s^\prime \sim M^* (\cdot|s, a)} \ls \lp \mu_{\theta} (s, a) - s^\prime  \rp^\top \Sigma_{\theta}^{-1} \lp \mu_{\theta} (s, a) - s^\prime  \rp + \log \lp \det \Sigma_\theta (s, a) \rp   \rs.
\end{align*}

Using the prediction model loss of either Eq.\eqref{eq:model_objective_mse} or Eq.\eqref{eq:model_objective_kl}, we can see that the model learning task has been transformed to be a supervised learning task. Any supervised learning technique can be employed to achieve efficient and effective model learning.

\subsubsection{Model Properties}

When a model has been obtained using the prediction model loss, we then care how well the model can help, in particular, how different a policy $\pi$ performs in the model and in the real environment, i.e., the value evaluation error,
$$
\|V^\pi_{M_\theta} - V^\pi_{M^*}\|_\infty.
$$

The \emph{simulation lemma} firstly proved in \citep{kearns02near} says that

\begin{thm}[Simulation Lemma]
\label{thm:simulation_lemma}
Given an MDP with reward upperbound $R_{\max}$ and transition model $M^*$, and a learned transition model $M_\theta$ with $\max _{s, a}\|M_\theta(s, a)-M^*(s, a)\|_{1} \leq \epsilon^{\max}_{m}$ and a learned reward function with $\max _{s, a}|R_\theta(s, a)-R(s, a)| \leq \epsilon_{r}$, the value evaluation error of any policy $\pi$ is bounded as 
\begin{equation}
	\|V^\pi_{M_\theta} - V^\pi_{M^*} \|_{\infty} \leq \frac{\gamma \epsilon^{\max}_{m} R_{\max }}{2(1-\gamma)^{2}} +\frac{\epsilon_{R}}{1-\gamma}.
\end{equation}
\end{thm}

The simulation lemma shows that the value loss corresponding to the model error $\epsilon^{\max}_{m}$ has a quadratic coefficient on the effective horizon $\frac{1}{1-\gamma}$. This means that the value loss grows quadratically fast as the horizon grows. When the reward function is out of the consideration, as discussed above, we can omit the $\epsilon_r$ term. Meanwhile, the simulation lemma also shows that the reward error is not severe compared to the model error.

We can also note the limitations of the simulation lemma. Compared with the learning loss Eq.\eqref{eq:model_objective_mse} where the data follows the distribution of the data-collecting policy, the model error $\epsilon^{\max}_{m}$ is measured as the maximum difference over all state-action pairs, which is not easily achieved or assessed in a practical way.

In the recent analysis \citep{slbo, mbpo, xu2020error}, the error of the learned transition model $M_\theta$ is measured over the data distribution, and thus directly connects with the learning loss Eq.\eqref{eq:model_objective_mse}. We present the result in Theorem \ref{thm:compounding_evaluation_loss}, named simulation lemma II. Note that we omit the reward function error since it is not essential.

\begin{thm}[Simulation Lemma II]
\label{thm:compounding_evaluation_loss}
Given an MDP with reward upper bound $R_{\max}$ and transition model $M^*$, and a data-collecting policy $\pi_D$, and a learned transition model $M_\theta$ with 
$$\mathbb{E}_{(s, a) \sim \rho^{M^*}_{\pi_D}} \left[D_{\mathrm{KL}} \bigl( M^*(\cdot|s,a), M_\theta (\cdot|s, a) \bigr)\right] \leq \epsilon^{\rho}_\textit{m},$$ 
for an arbitrary policy $\pi$ with bounded divergence,
$$\max_{s} D_{\textnormal{KL}} \bigl(\pi(\cdot | s), \pi_D(\cdot | s) \bigr) \leq \epsilon_\pi,$$
the value evaluation error of the policy is bounded as 
\begin{align}
    \vert V^{\pi}_{M_\theta} - V^{\pi}_{M^*} \vert \leq \frac{ \sqrt{2} R_{\textnormal{max}} \gamma}{(1-\gamma)^2} \sqrt{\epsilon_m} + \frac{2\sqrt{2} R_{\textnormal{max}}}{(1-\gamma)^2} \sqrt{\epsilon_\pi}.
\end{align}
\end{thm}
Note that we denote the distributional model error as $\epsilon^{\rho}_m$ to distinguish the uniform model error $\epsilon^{\max}_m$ in Theorem \ref{thm:simulation_lemma}.

The policy evaluation error of simulation lemma II contains two terms, the bias of the learned model, and the policy divergence between the evaluating policy $\pi$ and the data-collecting policy $\pi_D$. \cref{thm:compounding_evaluation_loss} indicates that the policy evaluation error w.r.t the model error $\epsilon^{\rho}_m$ also has a quadratic coefficient on the effective horizon $\frac{1}{1-\gamma}$, similar to \cref{thm:simulation_lemma}. The coefficient means a quadratic compounding error of learning in the model, which is the reason that studies such as \citep{mbpo} only adopt short rollouts, say, less than 10 steps, in the learned model. 

Moreover, \citet{xu2021error} provided a finite sample complexity result for the evaluation error by further relating the model error $\epsilon^{\rho}_m$ with the samples and model space of the model learning. Interested readers are referred to \citep[Corollary 2]{xu2021error} for the detailed analysis.

\subsubsection{Model Variants} 

Since the compounding error is due to the recursive state-action generation using the one-step transition model, a way of alleviating the issue is to predict many steps at a time. A multistep model \citep{asadi2019combating} takes the current state $s_t$ and a sequence of actions $(a_t, a_{t+1}, \ldots, a_{t+h})$ with length $h$ as the input, and predicts the future $h$ states. A (deterministic) multistep model is represented as 
$$
	(s_{t+1}, \ldots, s_{t+h}) = M_\theta^h (s_t, a_{t+1}, \ldots, a_{t+h}),
$$
which can also be trained by supervised learning. Intuitively, compared with the one-step model, the multistep model does not take the predicted \dquote{fake} states as the input and hence could avoid the compounding error within $h$ steps \citep{asadi2019combating}. Meanwhile, the dynamics across multi-steps can be much more complex than one-step dynamics, therefore, multistep transition predictions can have a larger error.

The transition models discussed before are all forward models, i.e., predict along time. In addition to the forward model, there also exist studies on the backward transition model in MBRL \citep{Edwards2018Forward, Goyal2019Recall, Lai2020Bidirectional,Lee2020Context, Wang2021offline}. The backward transition model takes the future state $s_{t+1}$ and action $a_t$ as inputs and predicts the state $s_t$. The backward model is often used to generate reverse data, which can help reduce the rollout horizon \citep{Lai2020Bidirectional} and could improve the sample efficiency of MBRL \citep{Wang2021offline}.

\subsection{Model Learning with Reduced Error}
\label{sec:model-imitation}

In the above model properties, we can see the horizon-squared compounding error is a major issue of model learning. The issue is mainly due to the use of prediction loss to learn an unconstrained model. 

\subsubsection{Model Learning with Lipschitz Continuity Constraint}

To reduce the compounding error, one way is to constrain the model. \cite{Venkatraman2015Improving} and ,\cite{Asadi2018LipschitzContinuityinMBRL} employed the Lipschitz continuity constraint for the models. They firstly employed Wasserstein distance \citep{vaserstein1969markov} to measure the similarity between two transition distributions. For any two distributions $P$ and $Q$ over space $\gX$, the Wasserstein distance between $P$ and $Q$ is defined as
\begin{align*}
    W \left(P, Q \right)=\inf _{\gamma \in \Pi\left(P, Q\right)} \mathbb{E}_{(x, y) \sim \gamma} \ls d(x, y) \rs.
\end{align*}
where $\Pi\left(P, Q\right)$ denotes the set of all joint distributions $\gamma (x, y)$, whose marginals are respectively $P$ and $Q$, and $d$ is the distance metric on $\gX$. Compared with other divergence measures (e.g., KL divergence and TV distance), Wasserstein distance can appropriately measure the similarity between two distributions with disjoint supports \citep{Asadi2018LipschitzContinuityinMBRL}.

When considering a state distribution $\rho$ over the state space $S$ rather than a single state $s$, we can also define the transition model, i.e., the generalized transition model, $M_\theta (\cdot|\rho, a)$ as
\begin{align*}
    M_\theta (s^\prime|\rho, a) := \int M_\theta (s^\prime | s, a) \rho (s) ds.
\end{align*}
$M_\theta$ is called $\epsilon_w$-accurate w.r.t. the Wasserstein distance if and only if $\sup_{s, a} W (M_\theta (\cdot|s, a), M^* (\cdot|s, a)) \leq \varepsilon_w$. Notice that $\epsilon_w$ measures the one-step error of the transition model and is connected to $\epsilon^{\max}_m$ in \cref{thm:simulation_lemma}. 

In MBRL, we desire an upper bound on the $n$-step error. For a fixed initial state distribution $\mu$ and a fixed sequence of actions $(a^0, \cdots, a^{n-1})$, the $n$-step error is defined as 
\begin{align*}
    & W (M^n_{\theta} (\cdot | \mu), M^{*,n} (\cdot | \mu)), \\
    & \text{where } M^n (s| \mu) := \sP \lp s_n = s| s_0 \sim \mu (\cdot), a_0 = a^0, s_1 \sim M (\cdot|s_0, a_0), a_1 = a^1, \cdots, a_{n-1} = a^{n-1}  \rp.
\end{align*}
The Lipschitz continuity is introduced for probabilistic transition models.
\begin{defn}
A probabilistic transition model $M$ is $K$-Lipschitz if and only if
\begin{align*}
    \sup_{a \in \gA} \sup_{\rho_1, \rho_2 \in \Delta (\gS)} \frac{W (M (\cdot|\rho_1, a), M(\cdot|\rho_2, a))}{W (\rho_1, \rho_2)} \leq K.
\end{align*}
\end{defn}

With a Lipschitz continuity constrained model, the $n$-step error can be bounded.
\begin{thm}[Theorem 1 in \citep{Asadi2018LipschitzContinuityinMBRL}]
\label{thm:n_step_error_lip_model}
Suppose the real transition model $M^*$ is $K^*$-Lipschitz and the learned transition model $M_\theta$ is $K$-Lipschitz and $\varepsilon_{w}$-accurate, then $\forall n \geq 1$,
\begin{align*}
    W (M^n_{\theta} (\cdot, \mu), M^{*,n} (\cdot, \mu)) \leq \varepsilon_w \sum_{i=1}^{n-1} (\widebar{K})^i,
\end{align*}
where $\widebar{K} = \min \{ K^*, K \}$.
\end{thm}

\citet{Asadi2018LipschitzContinuityinMBRL} further introduced two assumptions to simplify the analysis. One assumption is single action, i.e., $\gA = \{ a \}$. The other assumption is the state-only $K_R$-Lipschitz continuous reward function. Under the two assumptions, the value function under the transition $M$ is simplified as
\begin{align*}
    V_{M} (s) = \expect \ls \sum_{t=0}^\infty \gamma^t r (s_t) | s_0=s, a_t = a, s_{t+1} \sim M (\cdot| s_t, a_t), t = 0, 1,2, \cdots \rs.
\end{align*}
Then the value function error (of the policy taking the only action) can be bounded. 

\begin{thm}[Theorem 2 in \citep{Asadi2018LipschitzContinuityinMBRL}]
\label{thm:value_error_lip_model}
Under the same assumptions as in \cref{thm:n_step_error_lip_model}. Further, suppose that $\gA = \{a \}$ and the reward function only depends on the state and is $K_R$-Lipschitz continuous. Then $\forall s \in \gS$ and $\widebar{K} \in [0, \frac{1}{\gamma})$,
\begin{align}
    \labs  V_{M_\theta}(s) - V_{M^*}(s)  \rabs \leq \frac{\gamma K_{R} \epsilon_w}{(1-\gamma)(1-\gamma \widebar{K})},
\end{align}
where $\widebar{K} = \min \{ K^*, K \}$.
\end{thm}

Although the analysis is over-simplified, we can still notice that when $\widebar{K}$ is small, the evaluation error can be small. In other words, the compounding error can be controlled, compared with those in the simulation lemmas (i.e., \cref{thm:simulation_lemma} and \cref{thm:compounding_evaluation_loss}).

Since $\widebar{K}$ is the minimum between the Lipschitz constants of $M^*$ and $M_\theta$, the theorem suggests a trade-off on the Lipschitz constant of $M_\theta$. When $K$ is small and $K \leq K^\star$, $\widebar{K}$ is also small. Meanwhile, with a small $K$, the model might be hard to approximate $M^*$ with a large $K^*$ and thus $\epsilon_w$ could be large.



\subsubsection{Model Learning by Distribution Matching}
\label{sec:model-learning-gail}

The prediction loss employed in \cref{thm:simulation_lemma} and \cref{thm:compounding_evaluation_loss} minimizes the model error on each point of the state-action data. While the prediction loss minimization can be straightforwardly solved by supervised learning, the long-term effect of transitions is hard to be captured, resulting in the horizon-squared compounding error issue. Therefore, to learn the long-term effect of transitions, an idea is to match the distributions between the real trajectories and the trajectories rolled out in the learned model.

The idea of distribution matching has been employed in imitation learning through adversarial learning such as the GAIL method \citep{ho2016gail} that imitates the expert policy in an adversarial manner, where a discriminator $\mathfrak D$ learns to identity whether a state-action pair comes from the expert demonstrations and a generator $\pi$ imitates the expert policy by maximizing the
discriminator score. This corresponds to the minimax optimization problem $$
  \min_{\pi \in \Pi} \max_{\mathfrak D \in (0, 1)^{S \times A}} \mathbb{E}_{(s, a) \sim \rho_{\pi_E}}\left[\log \big(\mathfrak D(s, a) \big) \right] + \mathbb{E}_{(s, a) \sim \rho_{\pi}} \left[\log(1 - \mathfrak D(s, a) \big) \right],
$$
recalling that $\rho_\pi$ is the state-action distribution by running the policy $\pi$, and $\pi_E$ is the expert policy.  When the discriminator is optimal to the inner objective, i.e., $\mathfrak D^* (s,a) =  {\rho_{\pi_E} (s,a)}/\left({\rho_{\pi_E} (s,a) + \rho_{\pi} (s,a)}\right)$, the generator essentially minimizes the Jensen-Shannon (JS) divergence between $\rho_{\pi_E}$ and $\rho_{\pi}$ (up to a constant),
\begin{equation}
\label{equation:gail_js}
 \min_{\pi \in \Pi}  D_{\mathrm{JS}}(\rho_{\pi_E}, \rho_{\pi}) := \frac{1}{2} \left[ D_{\mathrm{KL}} (\rho_{\pi_E}, \frac{\rho_{\pi}+\rho_{\pi_E}}{2} )+D_{\mathrm{KL}} (\rho_{\pi}, \frac{\rho_{\pi}+\rho_{\pi_E}}{2}) \right],
\end{equation}
which achieves the goal of distribution matching and solves the compounding error issue of imitation learning theoretically \citep{zhang2020generative, wang2020computation, xu2020error, xu2021nearly} and empirically \citep{ho2016gail, ghasemipour2019divergence, ke2019imitation}.

Through the bridge of duel MDP \citep{zhang2017learning,shi2019taobao} that treats the environment also as an agent, the idea of distribution matching is brought for model learning \citep{shi2019taobao, Wu2019ModelImitation, xu2020error, Eysenbach2021Mismatched}. The transition model $M_\theta$, which takes the current state-action pair as inputs and predicts a distribution over the next state, is regarded as a policy in the imitation view. A discriminator is employed to distinguish expert state-action-next-state tuples from the ``fake'' ones. The minimax optimization problem in \citep{xu2020error} is then formulated as
\begin{align}
\label{eq:model_learning_gail}
    \min_{M_\theta} \max_{\mathfrak D} \mathbb{E}_{(s, a, s^\prime) \sim \mu^{M^*}}\left[\log \big(\mathfrak D(s, a, s^\prime) \big) \right] + \mathbb{E}_{(s, a, s^\prime) \sim \mu^{M_\theta}} \left[\log(1 - \mathfrak D(s, a, s^\prime) \big) \right],
\end{align}
where $\mu^{M^*}$ and $\mu^{M_\theta}$ are the joint state-action-next-state distributions induced by the data-collecting policy $\pi_D$ in the true environment $M^*$ and $M_\theta$, respectively. Formally, 
$$\mu^{M^*} (s, a, s^\prime) = \rho^{M^*}_{\pi_D} (s, a) M^* (s^\prime|s, a)\ , \ \mu^{M_\theta} (s, a, s^\prime) = \rho^{M_\theta}_{\pi_D} (s, a) M_\theta (s^\prime|s, a).$$
The optimal solution of $M_\theta$ minimizes the JS divergence between $\mu^{M^\star}$ and $\mu^{M_\theta}$, matching the trajectory distribution.
Different from \citep{xu2020error}, \citet{Wu2019ModelImitation} chose to minimize the Wasserstein distance between $\mu^{M^*}$ and $\mu^{M_\theta}$. In \citep{Wu2019ModelImitation, xu2020error}, they kept the data-collecting policy $\pi_D$ fixed during the training process of the transition model $M_\theta$. On the other hand, \citet{shi2019taobao} optimized the transition model and policy jointly, which forms a multi-agent adversarial imitation learning.

By the distribution matching, \citet{xu2020error} (Theorem 3) proved an improved  policy evaluation error bound as in \cref{theorem:gail_evaluation_loss}, which is named Simulation Lemma III.
\begin{thm}[Simulation Lemma III]
  \label{theorem:gail_evaluation_loss}
  Given an MDP with reward upper bound $R_{\max}$ and transition model with $M^*$, and a data-collecting policy $\pi_D$, and a learned transition model $M_\theta$ with 
  $$D_{\textnormal{JS}}(\mu^{M_\theta}, \mu^{M^*}) \leq \epsilon_{m}^{JS},$$
  for an arbitrary policy $\pi$ with bounded divergence, 
  $$\max_{s} D_{\textnormal{KL}} \bigl( \pi(\cdot | s), \pi_{\textit{D}}(\cdot | s) \bigr) \leq \epsilon_\pi,
  $$
  the policy evaluation error is bounded as
  \begin{equation}
  \vert V^{\pi}_{M_\theta} - V^{\pi}_{M^*} \vert \leq \frac{2 \sqrt{2} R_{\textnormal{max}}}{1-\gamma} \sqrt{\epsilon_m^{JS}} + \frac{2 \sqrt{2} R_{\textnormal{max}}}{(1-\gamma)^2} \sqrt{\epsilon_\pi}.  	
  \end{equation}
\end{thm}

We can see in \cref{theorem:gail_evaluation_loss} that the coefficient on the model error $\epsilon_m^{JS}$ is \emph{linear} w.r.t. the effective horizon, i.e., $\frac{1}{1-\gamma}$, which meets the lower bound \citep{xu2021error} and thus cannot be further improved in general. This means the compounding error issue is solved. 

One may further notice that the model error $\epsilon_m^{JS}$ is a different quantity from the error $\epsilon_m$ in \cref{thm:compounding_evaluation_loss}. This is true. To achieve the same value, the matching loss using the JS-divergence requires more samples than the prediction loss. In \citep{xu2021nearly}, the adversarial imitation approach has been improved to have a lower sample complexity than that of supervised learning.

\subsubsection{Robust Model Learning}

While in simulation lemma III the compounding error is reduced, the policy divergence term about $\epsilon_\pi$ can still be large. The policy divergence is the difference between the data-collecting policy and the target policy. In order to reduce the divergence, one direction is to use a data-collecting policy with a wide distribution.  \cite{zhang2021learning} proposed to use a distribution of policy to collect data, instead of a single policy, such that the divergence can be reduced. When using a distribution of the data-collecting policy, the model learning objective can be formulated as
\begin{equation}\label{eq:multi-task-gail}
\min_{M_\theta}
\mathbb{E}_{\pi \sim \mathcal{P}(\pi)} 
\big[ \max_{\mathfrak D_{\pi}}  \mathbb{E}_{(s,a,s')\sim \mu_\pi^{M^*}} [\log \mathfrak D_{\pi}(s,a,s')] + \mathbb{E}_{(s,a,s')\sim \mu_\pi^{M_\theta}} [\log(1-\mathfrak D_{\pi}(s,a,s'))] \big],
\end{equation}
where $\mu_\pi^{M_*}$ and $\mu_{\pi}^{M_\theta}$ are the joint state-action-next-state distributions. Empirically, the expectation operation $\mathbb{E}_{\pi \sim \mathcal{P}(\pi)} $ can be approximated via taking average over a set of sampled policies.

Furthermore, to particularly deal with concern-case interacting policies, i.e., the outlier policies with much different behavior from the majority, \cite{zhang2021learning} designed conditioned value at risk (CVaR) \citep{tamar2015optimizing} objective which focuses on the $\epsilon$-percentile of policies with the largest value discrepancy in simulation.

\subsection{Model Learning for Complex Environments Dynamics}


The mainstream realization of the environment dynamics model is an ensemble of Gaussian processes where the mean vector and covariance matrix for the distribution of the next state are built based on neural networks fed in the current state-action pair \citep{Chua2018PETS}. Such an architecture is shown to work well on MuJuCo robot locomotion environments, where the state observations are sufficient statistics for future derivation.
However, there still exist many complex environments which are hard to be directly modeled via the above method. Below we discuss two important aspects of complex environment dynamics modeling.

\textbf{Partial Observability.}
For partially observable environments, the observations may not be sufficient statistics for future derivation, which makes the environment a partially observable MDP (POMDP) \citep{spaan2012partially}. 
For model learning in a POMDP, belief state estimation is the classic solution, where an observation model $p(o_t|s_t)$ and a latent transition model $p(s_{t+1}|s_t,a_t)$ is learned via maximizing a posterior and the posterior distribution $p(s_t|o_1, \ldots, o_t)$ can be inferred.
With deep learning, the latent state distribution $p(s_t|o_1, \ldots, o_t)$ can be obtained via a recurrent neural network \citep{ha.worldmodel18}. \cite{hausknecht2015deep} further introduced a delta distribution for deterministic settings or a Gaussian distribution for stochastic settings.

\textbf{Representation Learning.}
For high-dimensional state space such as images, representation learning that learns informative latent state or action representation will much benefit the environment model building so as to improve the effectiveness and sample efficiency of model-based RL \citep{yang2021representation} on different aspects, including value prediction \citep{oh2017value} and model rollout \citep{nagabandi2018neural}. 
\cite{ha.worldmodel18} applied an autoencoder network to encode the latent state that can reconstruct the image.
\cite{hafer2020dreamer} proposed Dreamer to learn the latent dynamics for visual control tasks, in which an environment model (called world model) with visual encoder and latent dynamics is learned based on collected experience, and the model is shown to have the capability of performing long rollout and value estimation.
\cite{hafner2021mastering} further proposed  DreamerV2 that supports the agent to learn purely from the model rollout data and achieve human-level performance on 55 Atari game tasks. 
Compared with Dreamer,
DreamerV2 replaces the Gaussian latent as proposed in PlaNet \citep{hafner2019learning} with the discrete latent, which brings superior performance. The possible reason for such effects would be the discrete latent representation can better fit the aggregate posterior and handle multi-modal cases.
Considering the state-action pair distribution mismatch between the model training and model rollout stages, \cite{shen2020model} incorporated a domain adaptation objective into the model learning task to encourage the model to learn invariant representations of state-action pairs between the real data and rollout data.

\section{Model Usage and Integration with Model Learning}

\subsection{Planning with Model Simulation}
\label{sec_planning}

When a model is available, the most straightforward idea of utilizing the model is to plan in it. 
\emph{Planning} denotes any computational process that takes a model as input and produces or improves a policy for interacting with the modeled environment \citep{sutton2018reinforcement,moerland2020framework,moerland2020survey}. We will list the MBRL approaches that integrate planning into their methods or frameworks. We will categorize these approaches according to the planning methods they adopt. 

\noindent{\textbf{Model predictive control (MPC).}} MPC \citep{MPC_book} is a kind of model-based control method that plans an optimized sequence of actions in the model. Classical MPC approach requires to design a parametric model and policy, such as in the linear quadratic form, to fit the optimizer such as a quadratic optimization solver. Learning-based MPC \citep{learningMPC} has a tight connection with MBRL, particularly for nonlinear and black-box models.   In general, at each time step, MPC obtains an optimal action sequence by sampling multiple sequences and applying the first action of the sequence to the environment. Formally, at time step $t$, an MPC agent will seek an action sequence $a_{t:t+\tau}$ by optimizing:
\begin{equation}
\label{eq_mpc_basic}
    \max_{a_{t:t+\tau}}~\mathbb{E}_{s_{t'+1}\sim p(s_{t'+1}|s_{t'},a_{t'})}\left[ \sum_{t'=t}^{t+\tau}r(s_{t'}, a_{t'}) \right],
\end{equation}
where $\tau$ denotes the planning horizon. Then the agent will choose the first action $a_t$ from the action sequence and apply it to the environment. 

Black-box MPC regards Eq.(\ref{eq_mpc_basic}) as a black-box optimization problem and adopts some zero-order optimization methods to solve it. MB-MF~\citep{nagabandi2018mbmf} adopts a basic optimization method, i.e., the Monte Carlo (MC) method (also known as ``random shooting''), which samples a number of action sequences $a_{t:t+\tau}$ from the space of action sequence uniformly and randomly. By applying the action sequences in the model, the current state $s_t$ can be transited to $s_{t+\tau}$ following the transition distribution. The returns accumulated during the transition process are used to evaluate the action sequences. The action sequence with the highest evaluation will be preserved as the solution of Eq.(\ref{eq_mpc_basic}). The MC method is simple to implement and does not require many computational resources. However, because of the low efficiency of the random sampling process, it also suffers from high variance and could fail to sample a high reward action sequence when the action space is large. Recent advances in MPC methods focus on altering the sampling strategies~\citep{Chua2018PETS,hafner2019planet} and the sampling space~\citep{wang2020poplin}.

Replacing the MC method with CEM~\citep{botev2013cross}, PETS~\citep{Chua2018PETS}, and PlaNet~\citep{hafner2019planet} improves the optimization efficiency. Instead of sampling randomly and uniformly, CEM samples the action sequences from a multivariate normal distribution, which will be adjusted according to the evaluation of the sampled sequences such that the high-reward sequences can be sampled with a higher probability. This principle resembles many other derivative-free optimization methods \citep{nikolaus2016cmaes,yu2016racos,hu2017sracos}. As a result, other derivative-free optimization methods can also be used to solve Eq.(\ref{eq_mpc_basic}) and integrated into the MPC framework. In addition to altering the optimization method, POPLIN-A~\citep{wang2020poplin} further improves the optimization efficiency by altering the sampling space to a space of action bias. Specifically, POPLIN-A seeks a sequence of action residual to adjust an action sequence proposed by a policy. For example, POPLIN-A-Replan, a variant of POPLIN-A, alters Eq.(\ref{eq_mpc_basic}) to Eq~(\ref{eq_mpc_poplin_a_replan}) by introducing a policy function $\pi(s)$:
\begin{equation}
    \label{eq_mpc_poplin_a_replan}
    \max_{\delta_{t:t+\tau}}~\mathbb{E}_{s_{t'+1}\sim p(s_{t'+1}|s_{t'},\pi(s_{t'})+\delta_{t'})}\left[ \sum_{t'=t}^{t+\tau}r(s_{t'}, \pi(s_{t'})+\delta_{t'}) \right].
\end{equation}
The policy learns to imitate the MPC results. As a result, at the state the policy has met, it can propose a nearly optimal action sequence as an initial solution. Thus, POPLIN-A largely simplifies the optimization problem and improves planning efficiency.

With an imperfect model, the result of MPC could not be reliable. However, we can still make use of the planning result by integrating it into a model-free RL framework.   I2A~\citep{racaniere2017i2a} integrates Monte Carlo planning results into model-free RL framework  as the auxiliary information. Instead of applying the first action of the sequence with the highest return, I2A encodes several rollouts from the model to rollout embeddings. The embeddings will then be aggregated and used to augment the input of the model-free RL agent. As I2A does not rely on the planning results, it can successfully use imperfect models.

\noindent{\textbf{Monte Carlo tree search (MCTS). }}  
MCTS \citep{browne2012survey,chaslot2008mcts,silver2016alphago,silver2017alphazero} is an extension of Monte Carlo sampling methods. MCTS also aims at solving Eq.(\ref{eq_mpc_basic}). Unlike the MC methods mentioned in the MPC part, MCTS adopts a tree-search method. At each time step, MCTS incrementally extends a search tree from the current environment state~\citep{browne2012survey,chaslot2008mcts}. Each node in the tree corresponds to a state, which will be evaluated by some approximated value functions or the return obtained after rollouts in the model with a random policy~\citep{chaslot2008mcts} or a neural network policy~\citep{silver2016alphago,david2017general,silver2017alphazero}. 
Finally, action will be chosen such that the agent can be more likely transited to a state which has a higher evaluated value. In MCTS, models are generally used to generate the search tree and evaluate the state.  

AlphaGo~\citep{silver2016alphago} first uses MCTS to beat professional human players in the game of Go. AlphaGo first utilizes human expert data to pre-train a policy network, which is used to generate the search tree. For each decision timestep, AlphaGo uses MCTS to decide where to play the next stone on board. AlphaGo Zero \citep{silver2017alphazero} is able to defeat professional human players without any human knowledge. AlphaGo Zero trains a policy to mimic the planning outputs of MCTS, like POPLIN-A \citep{wang2020poplin}. The policy is then used to generate the search tree in MCTS. The procedure of AlphaGo Zero can be regarded as an iteration between improving a policy by planning and enhancing the planning by the policy. This training procedure has also been adopted by recent work to play the board game Hex \citep{anthony2017tree_search}.

Despite the conventional MCTS algorithm can only be used in discrete action space, \cite{coutoux2011cuct}and \cite{moerland2018alpha_zero} extended the MCTS framework to continuous action space by progressive widening \citep{coulom2007elo,chaslot2008progressive}, which adaptively determines the number of child actions of a state in the tree according to the total number of visits of the state. Further, when the true model is not provided, MCTS can be applied to a learned model. Value prediction network (VPN)~\citep{oh2017vpn} learns an abstract state transition model.
The abstract state transition model infers the next abstract state by taking the current abstract state and action as input, which is the same as the transition function in typical MDP.
However, the abstract state has no semantics of the corresponding state. The purpose of the abstract transition model is to transit the abstract state to an abstract state that can be used to make more precise value and reward predictions. As a result, given an action sequence, the VPN can predict the reward and the state value for the future states after taking the action sequence to the environment. VPN applies MCTS to the learned model to search an action sequence that has the highest bootstrapped environment return. With the abstract transition model, VPN can be applied to the tasks where the observation is the image, e.g. atari games. The experiments show that VPN can outperform DQN \citep{mnih2015human} in several atari games. With a similar framework, MuZero~\citep{schrittwieser2019muzero} further improves the performance in atari games. Muzero also learns a transition model but additionally learns an abstract policy, which outputs actions with the abstract states as inputs. Empirical results have shown huge advantages of Muzero in atari games, Go, chess and shogi.

\noindent{\textbf{Background planning. }} MPC and MCTS always begin and complete planning after the agent encounters a new state. This kind of method is also known as decision-time planning \citep{sutton2018reinforcement}. Another way of planning is \textit{background planning}, which uses the simulated data obtained from the model to improve the policy or value learning \citep{sutton2018reinforcement}. Dynamic programming \citep{sutton2018reinforcement}, tabular Dyna \citep{sutton1990dyna,sutton1991dyna}, and prioritized sweeping \citep{moore1993prioritized} all belong to background planning methods. Here, we introduce VIN \citep{tamar2016vin}, which implements a dynamic programming method, value iteration (VI), with a neural network. VIN reveals that the classic VI planning algorithm \citep{bellman1958dp} may be represented by a specific type of convolutional neural network (CNN). A step of VI can be achieved by passing a reward tensor to a CNN followed by a max-pooling layer. They embed such a module inside a standard feed-forward network and thus obtain a NN model that can learn to plan implicitly and provide the policy with useful planning results. 


\subsection{Data Augmentation with Model Simulation}


With a model, instead, we can generate any number of simulated samples as we want. We can use the simulated data for policy learning or value approximation. These kinds of integration of policy/value learning and models are known as Dyna-style methods. Dyna-style methods~\citep{Sutton90QPlanning} utilize the learned transition model to generate more experiences and then perform reinforcement learning on the dataset augmented by the model experiences. As a result, the models in Dyna-style methods are regarded as the data-augmenter for the policies. The main purpose of models is to generate simulated experiences for policy learning. In this sub-section, we will introduce value learning and policy learning with the simulated experience obtained from a model. 

\noindent{\textbf{Value estimation.}}
Monte Carlo (MC) value estimation \citep{tesauro1996mcsearch,sutton2018reinforcement} is the original method to approximate state values. 
Specifically, for an state $s_t$, it uses MC search to estimate $Q^\pi(s_t,a)$ by performing action $a$ in state $s_t$ and subsequently executing policy $\pi$ in all successor states. In MC search, many simulated trajectories starting from $(s_t,a)$ are generated following $\pi$. The value function $Q^\pi(s_t, a)$ will be estimated by averaging the cumulative reward of the trajectories. MC value estimation depicts an original model usage for value approximation, i.e., averaging the cumulative reward of the simulated trajectories. Another value approximation method is temporal-difference (TD) prediction \citep{tesauro1995td}, which is broadly used in many value-based RL methods \citep{lillicrap2015ddpg,mnih2015human}. In one step TD, the update target of the value of state $s_t$, $V(s_t)$, is determined by the value of the next state $V(s_{t+1})$ and the received reward $r_t$: $r_t + \gamma V(s_{t+1})$. Compared with the MC methods, one step TD does not need a environment model and thus is preferred by many model-free methods. The intermediate between MC methods and one-step TD is $H$-step TD, whose update target for $V(s_t)$ is $\sum_{t'=t}^{t+H-1}\gamma^{t'-t} r_{t'} + \gamma^{H}V(s_{t+H})$. \cite{Feinberg2018MVE} proposed Model-based Value Expansion (MVE), which indicates that $H$-step TD value prediction can reduce the value estimation error under some conditions, which is concluded in Theorem \ref{thm_mve}.

\begin{thm}[Theorem 3.1 in \citep{Feinberg2018MVE}]
\label{thm_mve}
Define $s_t$, $a_t$, $r_t$ to be the states, actions, and rewards resulting from following policy $\pi$ using the true dynamics $f$ starting at $s_0\sim v$ and analogously define $\hat{s}_t$, $\hat{a}_t$, $\hat{r}_t$ using the learned dynamics $\hat{f}$ in place of $f$. Let the reward function $r$ be $L_r$-Lipschitz and the value function $V^\pi$ be $L_V$-Lipschitz. Let $\epsilon$ be a be an upper bound 
\[
\mathop{\max}_{t\in[H]}~\mathbb{E}\left\Vert \hat{s}_t-s_t \right\Vert^2 \leq \epsilon^2,
\]
on the model risk for an $H$-step rollout. Then for any parameterized value function $\hat{V}$,  $H$-step model value expansion estimate $\hat{V}_H(s_t)=\sum_{t'=t}^{H+t-1} \gamma^{t'-t}\hat{r}_t' + \gamma^{H} \hat{V}(s_{t+H})$ satisfies
\[
\mathop{\text{MSE}}_{\nu}(\hat{V}_H) \leq c_1^2 \epsilon^2 + (1+c_2\epsilon)\gamma^{2H} \mathop{\text{MSE}}_{(\hat{f}^\pi)^{H}\nu}(\hat{V}),
\]
where $c_1$, $c_2$ grow at most linearly in $L_r$, $L_V$ and are independent of $H$ for $\gamma < 1$, $\mathop{\text{MSE}}_{\nu}(V)=\mathbb{E}_{S\sim \nu}\left[(V(S)-V^\pi(S)\right]$ measures the value estimation error, $(\hat{f}^\pi)^{H}\nu$ denotes the pushforward measure resulting from playing $\pi$ $H$ times starting from states in $\nu$, $\hat{V}$ is the . We assume $\mathop{\text{MSE}}_{(\hat{f}^\pi)^{H}\nu}(\hat{V}) \geq 2$ for simplicity of presentation, but an analogous result holds when the critic outperforms the model.
\end{thm}

Theorem \ref{thm_mve} implies that, if $\epsilon$ is small and the value function $\hat{V}$ is at least as accurate on imagined states $(\hat{f}^\pi)^H\nu$ as on those sampled from $\nu$:
\begin{equation}
\label{eq_mve_condition}
    \mathop{\text{MSE}}_{\nu}(\hat{V})\geq \mathop{\text{MSE}}_{(\hat{f}^\pi)^H\nu}(\hat{V}),
\end{equation}
the MSE of the $H$-step value estimation will approximately contract by $\gamma^{2H}$. Thus, the primary principle of MVE is forming $H$-step TD targets by unrolling the model dynamics for $H$ steps. Moreover, in order to satisfy Eq.(\ref{eq_mve_condition}), MVE trains the value function on the data sampled in the environment as well as the imagined data sampled in a learned model. However, MVE relies on fixed horizon $H$, which is task-specific and may change in different training phases and states space. 
Stochastic ensemble value expansion (STEVE) \citep{buckman2018steve} further improves MVE by interpolating between different horizons $H$ based on the uncertainty calculated using ensemble. 
It reweights the value targets of different depths according to their uncertainty, which is derived from both the value function and transition dynamics uncertainty. The rollout length with lower uncertainty will be assigned with a higher weight. 

\noindent{\textbf{Policy learning.}}
The data augmented by the model can also be used by model-free RL methods for policy improvement. The learned dynamic model can be regarded as a simulator where the policy can be trained.  \cite{kurutach2018metrpo} proposed Model Ensemble Trust Region Policy Optimization (ME-TRPO), which learns a policy via TRPO \citep{schulman2015trpo} from a set of learned models. The training process iterates between collecting data using the current policy, training the ensemble model with the environment data, and improving the policy in the ensemble model. Each model is learned from the real trajectories by minimizing the prediction error. When interacting with the ensemble model, in every step, ME-TRPO randomly chooses a model to predict the next state given the current state and action. The collected imagined trajectories are used to update the policy via TRPO. 
Further, \cite{slbo} studied such a learning framework from a theoretical perspective.
The authors find that if we establish a discrepancy bound 
\[
D_{\pi_\text{ref}}(\hat{M}) = L\cdot \mathbb{E}_{S_0,\dots,S_t\sim \pi_\text{ref},M^\star}\left[\lVert \hat{M}(S_t) - S_{t+1} \rVert \right],
\]
and update a model $\hat{M}$ and a policy $\pi$ following
\begin{equation}
    \label{eq_slbo_update}
    \pi, \hat{M} \leftarrow \mathop{\arg\max}_{\tilde{\pi}\in\Pi, \tilde{M}\in\mathcal{M}} ~V^{\tilde{\pi},\tilde{M}} - D_{\pi_\text{ref}}(\tilde{M}), ~~\text{s.t.}~~d(\tilde{\pi},\pi) \leq \delta,
\end{equation}
the policy performance in a true dynamical model $M^\star$ will improve monotonically. Here, $\pi_\text{ref}$ is a reference policy, $\hat{M}$ is the estimated model, $V^{\tilde{\pi},\tilde{M}}$ denotes the value of $\tilde{\pi}$ in model $\tilde{M}$, $\Pi$ is the policy space, $\mathcal{M}$ is the model space. Eq.(\ref{eq_slbo_update}) can be divided into two terms. Based on Eq.(\ref{eq_slbo_update}), as a result, \cite{slbo} proposed stochastic lower bound optimization (SLBO), which could be regarded as a variant of ME-TRPO. The authors discard the gradient of the first term w.r.t. the model for an approximation. As a result, the maximization of the first term is equal to updating the policy by an RL algorithm in the current model. The second term implies minimizing an $H$-step prediction of the model. The training procedure is quite similar to ME-TRPO but uses a multi-step L2-norm loss to train the dynamics. SLBO is built with theoretical guarantees but still has a problem. SLBO uses the model to roll out whole trajectories from the start state. However, due to the compounding error of the model, we may not rollout so long horizon. Model-based policy optimization (MBPO) \citep{mbpo}, on the other hand, samples the branched rollout in the model. MBPO begins a rollout from a state 
sampled in the real environment and runs $k$ steps according to policy $\pi$ and the learned model $p_\theta$. Moreover, MBPO also adopts Soft Actor-Critic \citep{haarnoja2018sac}, which is an off-policy RL algorithm, to update the policy with the mixed data from the real environment and learned model. MBPO also gives a monotonic improvement theorem with model bias and $k$-branch rollouts.

\begin{thm}[Theorem 4.3 in \citep{mbpo}]
\label{thm_mbpo}
Let the expected TV-distance between two transition distributions be bounded at each timestep by $\epsilon_m$, the policy divergence be bounded by $\epsilon_\pi$, the model error on the distribution of the current policy $\pi$ be bounded by $\epsilon_{m'}$, the true returns $\eta[\pi]$ and the returns from the $k$-branched rollout method satisfy:
\begin{equation}
    \label{eq_mbpo_bound}
    \begin{aligned}
    \eta[\pi] \geq \eta^\text{branch}[\pi] - C(\epsilon_m,\epsilon_\pi, \epsilon_{m'},k),~~
    C(\epsilon_m,\epsilon_\pi, \epsilon_{m'},k)  =
    2r_\text{max}\left[\frac{\gamma^{k+1}\epsilon_\pi}{(1-\gamma)^2}+\frac{\gamma^k\epsilon_\pi}{1-\gamma} + \frac{k}{1-\gamma}\epsilon_{m'} \right].
    \end{aligned}
\end{equation}
\end{thm}
Theorem \ref{thm_mbpo} implies if we can improve the returns under the model $\eta^\text{branch}[\pi]$ by more than $C(\epsilon_m,\epsilon_\pi, \epsilon_{m'},k)$, the policy improvement under the true returns can be guaranteed. Further the authors also prove that the optimal $k=\arg\min_{k}C(\epsilon_m,\epsilon_\pi, \epsilon_{m'},k) > 0$ for sufficiently low $\epsilon_{m'}$, which indicates that roll-out in the learned model with $k^\star$ steps is better than sampling full trajectories or discarding the model. Further, bidirectional model-based policy optimization (BMPO) \citep{Lai2020Bidirectional} adopts a backward model to reduce $C(\epsilon_m,\epsilon_\pi, \epsilon_{m'},k)$. BMPO introduces a backward dynamic model $q(s_t|s_{t+1},a_t)$ and a backward policy $\pi(a_t|s_{t+1})$ to accomplish backward trajectory sampling. Starting from a state $s_t$, BMPO will sample $k_1$ steps backward rollouts and $k_2$ steps forward rollouts. The policy will be learned from the mixture of the forward rollouts, backwards rollouts, and real environment data. The true returns of the policy learned by BMPO will be bounded by:
\[
\begin{aligned}
\eta[\pi] &\geq \eta^\text{branch}[\pi] - C^\text{BMPO}(\epsilon_m,\epsilon_\pi, \epsilon_{m'},k_1,k_2),\\
    C^\text{BMPO}(\epsilon_m,\epsilon_\pi, \epsilon_{m'},k_1,k2)  &=
2r_\text{max}\left[\frac{\gamma^{k_1+k_2+1}\epsilon_\pi}{(1-\gamma)^2}+\frac{\gamma^{k_1+k_2}\epsilon_\pi}{1-\gamma} + \frac{\max(k_1,k_2)}{1-\gamma}\epsilon_{m'} \right].
\end{aligned}
\]
Note that the bound of MBPO with the same rollout length to BMPO is $C(\epsilon_m,\epsilon_\pi, \epsilon_{m'},k_1+k_2)$. 
It is evident that BMPO obtains a tighter upper bound of the return discrepancy by employing bidirectional models. 

Typically, MBPO uses a short rollout length to avoid large compounding errors, which may limit the model usage. Recently, \cite{pan2020m2ac} proposed Masked Model-based Actor-Critic (M2AC), which can choose a longer rollout length to leverage the model better by discarding the samples with high uncertainty. M2AC computes the uncertainty of each sample by measuring the disagreement between one model versus the rest of the models in an ensemble model. M2AC simultaneously samples $B$ trajectories from the learned ensemble model. For each step, $B$ samples will be collected and sorted according to their uncertainty. Only the first $wB$ samples with the least uncertainty will be stored, others will be discarded. Experiment results demonstrate that M2AC can even benefit from longer model rollouts. On the contrary, the performance of MBPO drops rapidly as the rollouts become longer. 

Dyna-style methods can integrate model learning and model-free RL naturally. These methods have an impressive performance as well as a theoretical bound. As a result, Dyna-style algorithms attract lots of research interest in the MBRL community. A common and significant problem for these methods is how to tackle or alleviate the compounding errors. How to use the model to generate more reliable data and how to make better use of the imagined data are still open problems.

\subsection{Gradient Generation with White Box Model Simulation}
In the former sub-sections, we regard the dynamic model as a black box, with which we can transfer a state to another conditioned on an action. However, in many MBRL scenarios, the dynamic models are differentiable. A dynamic model could be a neural network \citep{heess15svg}, Gaussian process \citep{deisenroth2011pilco}, or a differentiable physics engine \citep{degrave2019differentiable}. We can utilize the internal structure of the models to facilitate policy learning. In this sub-section, we will introduce the approaches that use a white box dynamic model for policy learning. We list two categories of these approaches: differential planning and value gradient, both of which use the internal structure of the model to plan or learn a policy. 

\noindent{\textbf{Differential planning.}}
Planning in a white box model could be more data-efficient. The Monte Carlo trials discussed in Sec. \ref{sec_planning} can be altered by gradient-based search. In some cases, we can obtain the analytic form of the optimal policy for an MDP. Linear quadratic regulator (LQR) \citep{kwakernaak1969linear,todorov2005generalized} studies the MDP where the dynamic is linear, and the reward is quadratic. Particularly, the dynamic function is a linear function of state and action. Meanwhile, the reward function is a quadratic function of state and action. In this case, the optimal policy for each step is a linear function of the current state and can be derived from the parameters of the dynamic and reward functions. In the non-linear model, we can approximate the dynamic model to the first order and the cost function to the second order. LQR can then be applied to the approximated model, which is known as iterative LQR (iLQR) \citep{li2004ilqr,tassa2012ilqr}. As a result, we can linearize a learned dynamic model and use iLQR to determine the approximately optimal action at each step \citep{watter2015e2c,levine2013gps,levine2014gps_withmodel}. Guided Policy Search (GPS) \citep{levine2013gps} uses iLQR to draw samples from a white-box model. The samples are used in two ways: producing an initial neural network policy by BC and updating the policy via policy gradient. GPS is a successful integration of planning and reinforcement learning. The GPS framework has been generalized to the case where the dynamic model is unknown \citep{levine2014gps_withmodel}, or the input is high-dimensional images \citep{levine2015gps_image,levine2016gps_image}. In recent work, \cite{zhang2019solar} proposed stochastic optimal control with latent representations (SOLAR) based on the GPS framework. SOLAR defeats an MPC method \citep{ebert2018visual} in the task of learning image-based control policy on a real robot. This result indicates the promising potential of GPS for solving real-world tasks.

Another way of differential planning is utilizing the gradient of the dynamic model for action sequences search. Eq.(\ref{eq_mpc_basic}) can be optimized by gradient descent methods if the dynamic model is differentiable. \cite{srinivas2018upn} proposed universal planning networks (UPN), which involves a gradient descent planner (GDP). The GDP uses gradient descent to optimize an action sequence to reach a goal. In UPN, the reward is related to the distance between the final state $s_{t+\tau}$ and a goal state $s_g$. The reward is given at the last step of the planned rollout. Thus, in UPN, Eq.(\ref{eq_mpc_basic}) is instantiated by 
\begin{equation}
\label{eq_upn}
\min_{a_{t:t+\tau}}~\mathbb{E}_{s_{t'+1}=f(s_{t'},a_{t'})}\lVert s_{t+\tau} - s_g \rVert_2^2,
\end{equation}
where $f(s_{t'},a_{t'})$ is the dynamic model. 
In fact, we can write $s_{t+\tau}$ as iteratively calling the dynamic model:
\begin{equation}
    \label{eq_iterative_call}
    s_{t+\tau} = f(\dots f(f(f(s_{t}, a_{t}), a_{t+1}), a_{t+2}), \dots, a_{t+\tau-1}).
\end{equation}
As $f(s_{t'},a_{t'})$ is differentiable, the gradient of Eq.(\ref{eq_upn}) can be passed to $a_{t:t+\tau-1}$ through the dynamic model. GDP alters the Monte Carlo search in MPC to gradient-based search, which is more data-efficient. Despite the high data efficiency, gradient-based methods are prone to sticking to local minimums. However, sample-based methods, e.g. CEM, do not suffer from this problem. \cite{bharadhwaj2020cem_gradient} integrated CEM and gradient-based search to optimize the action sequence. For each CEM iteration, they will use a gradient-based search to refine the action sequences sampled by CEM before the sequences are evaluated. Their experiments show that their method can also avoid the local minima. Compared with CEM, their methods can converge faster and obtain better or equal performance.

\noindent{\textbf{Value gradient.}}
The policy gradient can also be passed through a white-box model. Probabilistic inference for learning control (PILCO) \citep{deisenroth2011pilco} models the dynamic model by the Gaussian process \citep{seeger2004dgml}. PILCO contains four stages, i) learns a probabilistic Gaussian process dynamic model from data; ii) evaluates the policy via approximate inference in the learned model; iii) obtains the gradient of the policy w.r.t. the policy evaluation; iv) updates the policy parameters to maximize the policy evaluation via conjugate gradient or L-BFGS \citep{peter2006cg}. The training process iterates between collecting data using the current policy and improving the policy. Although PILCO has a graceful mathematics form for the policy gradient and is able to estimate the model uncertainty with by Gaussian process naturally, its GP model is hard to scale in high-dimensional environments. To improve the scalability of PILCO, \cite{gal2016improving} replaced the GP with a Bayesian neural network to model the dynamic \citep{mackay1992bayesian}. They scale PILCO to high dimensions as well as retain the probabilistic nature of the GP model. 

For the dynamic model instantiated by a common neural network, e.g. fully-connected or convolutional neural network, the policy gradient can be estimated by backpropagating through the learned model \citep{mohamed2020mcg}. These methods typically involve two parts: re-parameterization of distributions \citep{kingma2014vae,rezendemw2014reparam} and policy gradient backpropagation through a model. We will take stochastic value gradient (SVG) \citep{heess15svg} as an example. SVG re-parameterizes the policy $\pi(a|s)$ by regarding the policy as deterministic function of state $s$ and a noise $\eta$: $a=\pi(s,\eta;\theta)$, where $\eta\in \rho(\eta)$ is a random vector, $\theta$ is the parameter of $\pi$. Similarly, the dynamic model is $s'=f(s,a,\xi;
\phi)$, where $\xi \in \rho(\xi)$ is a random vector, $\phi$ is the parameter of $f$. For any state-action sequence $(s_t,a_t,s_{t+1},a_{t+1},\dots)$, we can infer the corresponding $(\eta_t,\xi_t,\eta_{t+1},\xi_{t+1},\dots)$ with $\pi(s,\eta)$ and $f(s,a,\xi)$, such that $\pi(s_{t'}, \eta_{t'};\theta)=a_{t'}$, $f(s_{t'},a_{t'}, \xi_{t'};\phi)=s_{t'+1}$, $\forall t'\in\{t,t+1,\dots\}$. Further, the value function $V^\pi(s_t)$ can be estimated by $H$-step return plus a parameterized value function $v(\cdot)$:
\begin{equation}
    \label{eq_svg_value}
    \begin{aligned}
    V^\pi(s_t)=\sum_{t'=t}^{t+H-1}&\left[\gamma^{t'-t}r(s_{t'}, a_{t'}) + \gamma^H v(s_{t+H})\right],~~a_{t'}=\pi(s_{t'}, \eta_{t'};\theta),\\
    s_{t'+1}=f(s_{t'}, a_{t'}, \xi_{t'};\phi) &=f(f(s_{t'-1}, a_{t'-1}, \xi_{t'-1};\phi), \pi(s_{t'}, \eta_{t'};\theta), \xi_{t'}) = \cdots
    \end{aligned}
\end{equation}
Note that each state $s_{t'}$ can be obtained by recursively calling the dynamic function like Eq.(\ref{eq_iterative_call}). As a result, we can write the $V^\pi(s_t)$ as a function of $s_t, (\eta_t,\xi_t,\eta_{t+1},\xi_{t+1},\dots)$, parameterized by $\theta, \phi$:
\begin{equation}
    \label{eq_svg_larget_value}
    V^\pi(s_t) = F\left(s_t,\eta_t,\xi_t,\eta_{t+1},\xi_{t+1},\dots;\theta,\phi\right).
\end{equation}
As the objective of RL is maximizing $V^\pi(s_t)$, the policy gradient at $s_t$ is $-\nabla_\theta V^\pi(s_t)$. In SVG($1$), SVG($\infty$) \citep{heess15svg}, and SVG-$H$ \citep{amos2021svgh}, the value function is estimated with $1$, $\infty$, and $H$-step return. 
Moreover, the estimation is based on real-world trajectories. It will introduce a likelihood ratio term for the model predictions and increase the variance of the gradient estimate.
Model-augmented actor-critic (MAAC) \citep{clavera2020maac}, dreamer \citep{hafer2020dreamer}, and imagined value gradients (IVG) \citep{byravan2019ivg}, instead, entirely rely on the predictions of the model, removing the need for likelihood ratio terms. Particularly, to estimate the action or state values, MAAC  will sample an $H$-step rollout in the model with the current policy. Additionally, MAAC estimates the action-value function, i.e. Q-function $Q^\pi(s,a)$, rather than the state value function $V^\pi(s)$. It has been proved that the gradient error of MAAC can be bounded by model and Q-function errors. The authors further give the lower bounds of improvement for MAAC in terms of model error and function error, which implies that the policy can be improved monotonically with a precise dynamic model and an accurate Q-function. 

As neural network models are broadly used for model learning, the white box model-based methods can be naturally applied to these models. These methods can directly calculate the gradient of the RL objective w.r.t. the policy of action sequences. As a result, these methods have the potential to alter the trials-and-errors exploration and policy improvement paradigm to gradient descent. However, these methods currently suffer from gradient bias because of the model error. How reduce the gradient bias and variance is a future direction of these kinds of methods.

\subsection{Value-aware and Policy-aware Model Learning}
Previous MBRL works mainly treat model learning and model usage separately, which may result in a mismatch of learning objectives between models and policies. That is, the model is trained to give accurate predictions on all training data, while the policy is optimized to achieve high performance in the true environment. Therefore, a model with a small prediction error on the training dataset does not always imply a policy with high rewards \citep{lambert2020objective}. To this end, \citet{farahmand2017value} proposed the value-aware model learning (VAML) framework to address this problem by incorporating value function information into model learning. Intuitively, when the model-generated data is used to update value functions, we should focus more on the estimation error of the value target rather than the error of the next state. To be more specific, VAML optimizes the model to minimize the one-step value estimation difference between using environment and model:
\begin{equation}
    \mathcal{L}_{V}(\hat{p}, p, \mu)=\int \mu(s, a)\left|\int p\left(s^{\prime} \mid s, a\right) V\left(s^{\prime}\right) \mathrm{d} s^{\prime}-\int \hat{p}\left(s^{\prime} \mid s, a\right) V\left(s^{\prime}\right) \mathrm{d} s^{\prime}\right|^{2} \mathrm{~d}(s, a)
\end{equation}
By minimizing the above loss, the bootstrap target calculated using the model will be close to that using the true environment. \citet{voelcker2021value} further improved VAML by taking the gradient of value function into account, and proposed to learn a model that is more accurate on the state-action pairs with large value function gradients.
Similarly, for policy gradient based algorithm \citep{abachi2020policy}, the model should provide accurate estimate of policy gradient:
\begin{equation}
\nabla_{\theta} \hat{J}(\theta)=\mathbb{E}_{(s, a) \sim \hat{\rho}_{\pi_{\theta}}}\left[Q^{\pi_{\theta}}(s, a) \nabla_{\theta} \log \pi_{\theta}(a \mid s)\right],
\end{equation}
 and can be optimized by minimizing the difference between
$\nabla_{\theta} \hat{J}(\theta)$ and the ground-truth $\nabla_{\theta} J(\theta)$. By integrating value or policy information into model learning, the learned model can be more suitable for current updates and more robust than traditional maximum likelihood methods, especially when the model capacity is insufficient to fully represent the environment \citep{voelcker2021value}.

\section{Model-based Methods in Other Forms of RL}

\subsection{Offline RL}


Offline reinforcement learning studies the methodologies that enable the agent to directly learn an effective policy from an offline experience dataset without any interaction with the environment dynamics \citep{levine2020offline}. 
In general, given a collected experience dataset $\mathcal{D}=\{(s, a,r,s')\}$, the whole offline RL processing can be framed as
\begin{equation}
    \min_\pi \mathcal{L}(\mathcal{D}, \pi),\label{eq:offlinerl}
\end{equation}
where the design of the loss function $\mathcal{L}$ is the focus of different offline RL works. 
With such a setting, the agent will not require to interact with the environment before a satisfying policy has been learned. Thus, offline RL techniques can be applied to a much wider range of real-world applications.

Despite the non-interaction nature of offline RL makes its training objective  Eq.\eqref{eq:offlinerl} similar to that of supervised learning,
the key challenge of offline RL is the extrapolation error, also studied as an out-of-distribution (OOD) problem from the data perspective, caused by the discrepancy between the underlying behavior policy generating the dataset and the current learning policy \citep{kumar2020conservative}.


Model-free offline RL mainly designs algorithms that are  constrained by the offline dataset to avoid extrapolation errors  \citep{fujimoto2019off,kumar2020conservative,peng2019advantage,chen2020bail} without using extra data. As a result, the learned policies are usually conservative since the dataset itself always limits the appropriate generalization of the learning policy beyond the offline dataset \citep{Wang2021offline}. 

Model-based offline RL, on the other hand, first builds an environment model based on the offline dataset and then trains the policy based on the model (and the data). The key advantage of building the environment model in offline RL is to leverage the model's generalization ability to perform a certain level of exploration and also to generate additional training data to improve the policy performance. 

Nevertheless, since the offline data is often quite limited, the learned model is considered untrustable most methods take a conservative strategy. 
MORel \citep{kidambi2020morel} constructs a pessimistic MDP (P-MDP) using the offline dataset, where the state transition model is additionally trained and used for detecting whether the current state-action pair is OOD. If OOD is detected, the model will transit to a terminal state and outputs a negative reward. As such, the agent in the P-MDP tends to learn to avoid OOD situations and thus reduce the extrapolation error in the learning process.
MOPO \citep{yu2020mopo} derives a policy value lower bound based on a learned model and incorporates a penalty term into the reward function based on the model uncertainty so as to discourage the agent from entering the OOD region.
To bypass the error caused by regarding uncertainty estimated via the deep neural networks as the guidance of avoiding OOD problems, COMBO \citep{yu2021combo} trains a value function based on both the offline data and the rollout data generated by the learned model. Moreover, the value function is regularized on the OOD data generated via model rollout. 

Other than the conservative strategies that avoid entering into OOD regions, it is possible to generalize. MAPLE~\citep{chen2021offline} was the first to borrow the generalization ability of meta RL for offline RL. It derives an adaptable policy through a meta RL method to make adaptive decisions directly in OOD regions, which provided an alternative way to deal with OOD data. 

\subsection{Goal-conditioned RL}
Goal-conditioned reinforcement learning (GCRL), also named as goal-oriented RL \citep{liu2022goal}, deals with the tasks where agents are expected to achieve different goals \citep{pitis2020maximum} in an environment or complete a complex task via achieving a series of goals. In GCRL, the observation (or the state) of the agent is usually augmented with a goal, which is normally represented as a mapped vector $g\in \mathcal{G}$ from a target state, i.e., $g=\phi(s_\text{target})$. The mapping function $\phi: \mathcal{S} \mapsto \mathcal{G}$ can be designed based on specific tasks or just the identity function.
The resulted goal can be regarded as being sampled from a distribution $p_g$.
Then, the reward function is defined based on the (state, action, goal) tuple as $r: \mathcal{S}\times \mathcal{A} \times \mathcal{G} \mapsto \mathbb{R}$. In such a setting, the agent policy $\pi: S\times\mathcal{G} \mapsto \Omega(\mathcal{A})$ 
is trained to maximize the expected goal-conditioned return as
\begin{equation}
    \max_\pi J(\pi) = \mathbb{E}_{p, p_g, \pi} \Big[\sum_t \gamma^t r(s_t, a_t, g) \Big] ~.
\end{equation}

To efficiently train the agent to achieve various goals, \emph{goal relabeling} techniques are widely used. In hindsight, experience replay (HER) \citep{andrychowicz2017hindsight}, the goal is relabeled from the trajectories where the original goal may not be achieved, which is motivated by learning from failure cases. Specifically, in HER, the relabeled goals are built via mapping a randomly picked state in a trajectory of the replay buffer. Other following works seek different ways to generate the goal, including using GANs to generate goals with different difficulty scores \citep{florensa2018automatic}, planning the goals with search techniques based on experience data \citep{lai2020hindsight,eysenbach2019search}, etc.


To further enhance the diversity of the generated goals, thus the robustness and effectiveness of the trained goal-conditioned policy, recent attempts have been made for goal planning via model-based methods \citep{nair2019hierarchical} studied visual prediction models to build environment dynamics based on visual observations to enable subgoal generation and planning for robot manipulation tasks, which demonstrates substantial performance gain over the baseline model-free RL methods and planning methods without subgoals. 
\cite{zhu2021mapgo} proposed foresight goal inference (FGI) method to plan goals based on a learned environment dynamics model, then the simulated trajectories are generated based on the goal, goal-conditional policy, and the dynamics model. With model-based RL methods, the training scheme achieves superior sample efficiency than model-free GCRL baselines.

With the high-fidelity environment dynamics model, it is promising to leverage various techniques such as graph search, model inference, and heuristics created by human domain knowledge to generate highly useful goals to train the goal-conditioned policy and guide the policy's decision-making at the inference stage.

\subsection{Multi-agent RL}\label{sec:marl}
Multi-agent reinforcement learning (MARL) studies the sequential interaction strategies among a set of agents $i=1,2,\ldots,n$ in the  environment, where each agent $i$ is self-interested and aims to maximize its own payoff in terms of expected return as
\begin{align}
    \max_{\pi^i} J^i(\pi^{i}, \pi^{-i}) = \mathbb{E}_{p,\pi^{i},\pi^{-i}}
    \Big[\sum_{t=0}^\infty \gamma^{t} r^{i}(s_{t}, a_{t}^{i}, a_{t}^{-i}) \Big]~,
\end{align}
where the state transition dynamics $p: \mathcal{S}\times \mathcal{A}^1 \times \cdots \times \mathcal{A}^n \mapsto \Omega(\mathcal{S})$ now depends on the joint action $(a_t^i, a_t^{-i})$ from all the interacting agents, $\mathcal{A}^i$ denotes the action space of agent $i$, $r^{i}:  \mathcal{S}\times \mathcal{A}^1 \times \cdots \times \mathcal{A}^n \mapsto \mathbb{R}$ denotes the reward function for agent $i$. 

Different from single-agent RL, an extra dynamics when seeking the solution in MARL comes from the non-stationarity of the multi-agent game \citep{papoudakis2019dealing}. Ideally, in a Markov game, the Markov perfect equilibrium (MPE) is a profile of policies of the participating agents, where each agent $i$ has no incentive to change its current policy $\pi^i$ since for each state $s$, $J^i(\pi^i, \pi^{-i}) \geq J^i(\tilde{\pi}^i, \pi^{-i}), \forall \tilde{\pi}^i$ \citep{fink1964equilibrium}. Ideally, the MPE is the solution sought by MARL algorithms, while an approximate $\alpha$-MPE means in such a profile of policies, every agent achieves the value with the gap no larger than $\alpha$ compared with the value of its MPE policy \citep{subramanian2021robustness}. 

In recent work, \cite{subramanian2021robustness} performed an early theoretic analysis on model-based MARL. Particularly, they first proved that if a Markov game (i.e., the real multi-agent environment) is approximated by another game (i.e., the model of the multi-agent environment), then the MPEs from these two games are close to each other. Then, they derived that $\tilde{O}(|\mathcal{S}||\mathcal{A}|(1-\gamma)^{-2}\alpha^{-2})$ samples are sufficient to achieve an $\alpha$-MPE with high probability.
More specifically, for the two-agent zero-sum Markov game, \cite{zhang2020model} derived the sample complexity of the model-based MARL methods and showed that it is lower than the complexity of model-free MARL methods as derived in previous work \citep{bai2020provable}.


From the perspective of one agent, the environment it interacts with consists of the opponent agents and the environment dynamics which transits the state according to the joint actions of all agents. As such, the task of learning the multi-agent environment can be decoupled as learning opponent models and the environment dynamics. Opponent modeling \citep{he2016opponent} is a well-studied topic in multi-agent RL, while the work on environment dynamics learning in multi-agent RL is rare. \cite{mahajan2021tesseract} studied the problem of dimension explosion issue of the action space caused by the number agents and proposed model-based Tesseract, where Bellman equation is built is a tensorized form while the reward function and state transition dynamics are realized by low-rank Canonical-Polyadic decomposition. The theoretic analysis shows that the model-based Tesseract achieves an exponential gain in sample efficiency of $O(|\mathcal{A}|^{n/2})$.

From the analysis of \citep{zhang2021model}, the sample efficiency of MARL can be decomposed into two parts, i.e., dynamics sample complexity, which measures the amount of interactions with the real environment, and the opponent sample complexity, which measures the amount of interactions between the ago agent $i$ and other agents $\{-i\}$. With this regard, it is natural to derive the value discrepancy of the agent policy in the multi-agent environment model (i.e., with the state dynamics model and the opponent models) and the real environment with respect to the error terms of the state dynamics model and opponent models when training the policy in via Dyna-style model rollout. The bound shows that the opponent models with higher modeling error contribute larger to the discrepancy bound, which motivates the design of the algorithm called adaptive opponent-wise rollout policy optimization (AORPO) \citep{zhang2021model}. Specifically, the rollout scheme of AORPO allows the opponent models with lower generalization errors to sample longer trajectories while the shorter sampled trajectories can be supplemented with a communication protocol with real opponents. 
\cite{kim2020communication} proposed a communication mechanism called intention sharing (IS), where each agent builds the environment dynamics and opponent models and generates the rollout trajectory. Then a compressed representation is learned from the rollout trajectory to carry the intention, which is sent as a message to other agents for better coordination.

As stated in a recent survey on model-based MARL \citep{wang2022model}, the research in this direction has just started with only a few established works. The potential topics that are promising for the future development of model-based MARL include improving the scalability of centralized methods, and the new design of decentralized methods and communication protocols based on the learned models. 


\subsection{Meta RL}


Meta reinforcement learning~\citep{duan2016rl2,houthooft2018epg,yu2018shallow} studies the methodologies that enable the agent to generalize across different tasks with few-shot samples in the target tasks. In this process, we have a set of tasks for policy training, but the deployed tasks are unknown, can be OOD compared with the distribution of the training tasks~\citep{Lee2020Context}, and even can be varied when deployed~\citep{luo2022aaai}. Tasks have different definitions in different scenarios, e.g., differences in reward functions~\citep{finn2017,rothfuss2019promp}, or parameters of dynamics~\citep{peng2018dynamics_randimization,zhang2018ijcai}.
The challenge is to design efficient mechanisms to extract suitable information for the tasks and adjust the behavior of the policy based on the information.

Model-based meta RL methods learn dynamics models from the training tasks, adapt the dynamics models to the target tasks via few-shot samples, and finally generate actions via model predictive control (MPC) \citep{mpc} algorithms or a meta-policy (also called contextual policy) trained with RL methods~\citep{context_aware}.
\citet{learn_to_adapt} considered the task distribution is non-stationary when deployed, e.g., legged robots might lose a leg, face to novel
terrains and slopes when deployed. It solves the problem by learning an adaptive predictive model $\hat p_{\theta'} (s_{t+1}|s_t,a_t)$  with parameters $\theta'$, where $\theta'=u_{\phi}(\mathcal{D}_{test}, \theta)$ corresponds to model parameters that were updated using an adapter $u_\phi$ parameterized by $\phi$ and the collected dataset $\mathcal{D}_{test}$. $\phi$ and $\theta$ are trained to use the passed $M$-step data points to compute an optimal $\theta'$ which will 
minimize the negative log-likelihood $\mathcal{L}$ of future $K$-step data points:
\begin{align*}
    \min_{\theta, \phi} \mathbb{E}_{\tau(t-M, t+K) \sim \mathcal{D}}\left[\mathcal{L}(\tau(t, t+K), \theta') \right],~~s.t., \theta' = u_{\phi}(\tau(t-M, t-1),\theta),
\end{align*}
where $\tau(t-M, t+K) \sim \mathcal{D}$ corresponds to trajectory segments from $t-M$ timestep to $t+K$ timestep sampled from our previous experience.
It implements two adapters, i.e., gradient-based adaptive learner (GrBAL), which uses a gradient-based meta-learning as in \citep{maml} to perform online adaptation, and recurrence-based adaptive learner (ReBAL), which utilizes a recurrent model to learn to adapt via the gradient of $\mathcal{L}$. When deployed, it runs an MPC to generate actions using the adapted model $\hat p_{\theta'}$.   
MOLe \citep{continu_adapt} extends the approach to lifelong learning scenarios, i.e., continual online learning from an incoming stream of data. 
MOLe develops and maintains a mixture of models to handle non-stationary task distributions.  The mixture of models is a set of dynamics models split by a task indicator $\hat p_{\theta(T_i)}(s_{t+1}|s_t,a_t)$, where $T_i$ denotes a task. It designs an expectation-maximization (EM) algorithm to update all of the model parameters. When deployed, new models will be instantiated for task changes, and old models will be recalled when previously seen tasks are encountered again. CaDM \citep{context_aware} defines the tasks via the dynamics parameters $c$ (e.g., friction) and aims to learn a forward dynamics model from a set of training environments with contexts sampled
from $p_{train}(c)$ that can produce accurate predictions for test environments with unseen contexts sampled from $p_{test}(c)$. CaDM solves the problem via learning a context-aware dynamics model $\hat p_{\theta}(s_{t+1}|s_t,a_t, z_t)$, where $z_t=u_{\phi}(\tau(t-M, t-1))$, which is similar to the framework of ReBAL. CaDM proposes three auxiliary tasks, including forward prediction, backward prediction, and future-step prediction, to regularize the representation of $z$ for better generalization ability. 
\cite{meta_flight} considered a specific application that a policy should control a flight with suspended unknown payloads. As a solution, a context-aware dynamics model $\hat p_{\theta}(s_{t+1}|s_t,a_t, z_t)$ is also constructed to be aware of different payloads. \citet{meta_flight} inferred the context via a Gaussian with diagonal covariance $\mathcal{N}(\mu_t, \Sigma_t) \approx p_{\phi}(z\mid \tau_{:t-1})$ and formulated the meta-objective based on variational inference~\citep{kingma2014vae}.

The basic idea of learning to adapt has been used to solve the dynamics gap, also called reality gap, between training and test in many real-world applications. \cite{openaicube} taught an adaptive controller to solve a Rubik’s cube with a humanoid robot hand in a real-world environment with disturbances. 
\citet{anymal} learned an adaptive controller for quadrupedal robots which can make robust perceptive locomotion in an assortment of challenging natural and urban environments over different seasons. All of these works rely on a latent state $z$ to achieve adaptation. However, the generalization ability of $z$ itself is seldom considered, which can be further investigated in future work. Recent applications of model-based meta RL also have shown its generalization ability in complex tasks~\citep{cau_adapt,cau_adapt2}. These works also show that model-based meta RL with MPC is easy to introduce extra constraints of safety to action generation, which can be a potential advantage of model-based meta RL compared with model-free methods.

In particular, when the agents have access to a simulator and deal with tasks from both simulation and reality, it relates to Sim2Real \citep{peng2018dynamics_randimization, yu2017preparing, tan2018sim, rusu2017sim,codas}, which focuses on how to transfer a policy trained in the simulator to the real world. In such a case, model learning can be easier by making use of the off-the-shelf simulator. For example, \citet{golemo2018sim} proposed training a neural network to predict the difference between real data and simulated data. The output of the network was then used to compensate for the unreal parts of the simulator. Experiments have shown this neural-augmented method can be more efficient than directly learning a model to predict the next state. In vision-based RL, \citet{codas} trained a mapping function to align the representation space of high-dimensional observations in the real environment to low-dimensional state space in the simulator, then a policy can be trained in the simulator and deployed in the real world directly, by mapping observed images from the real world to aligned states and feeding the inferred states into the trained policy.
Moreover, \citet{hwangbo2019learning} trained an actuator network to produce reasonable predictions for the actuator of real robots since the actuator model of the simulator is not accurate enough.
Furthermore, the SimGAN framework \citep{jiang2021simgan} utilized GAN \citep{GAN} to generate the simulation parameters, e.g., actual motor forces or contact forces, which were then used to replace the corresponding components of the simulator. As a result, the simulated data can be closer to the real data compared with an original simulator or fully learned dynamics models. In more complicated real-world scenarios, learning an accurate model can be a big challenge due to the high-dimensional state-action space and complex interactions between different objects. 

In the future, it is tempting to build more realistic hybrid models combining analytical simulators and neural networks, which can seamlessly take advantage of innovations in both fields.

\subsection{Automated Methods on Model Learning and Usage}

Compared to MFRL, the design and tuning of MBRL methods tend to require more human effort due to their complex algorithm procedure and sensitivity to different hyperparameters. Take Dyna-style methods as an example. Besides designing the model-free counterpart, Dyna-style methods should also consider alternate optimization of model and policy, model planning steps, and the ratio of simulated data to real data \citep{lai2021effective}. To this end, automated MBRL methods have been investigated to automate the MBRL pipeline and search for better hyperparameters. For example, The reinforcement on reinforcement (RoR) framework proposed in \citet{dong2020intelligent} additionally trained a high-level controller through DQN \citep{mnih2013playing} to control the sampling and training process of a Dyna-style MBRL method. \citet{zhang2021importance} utilized population-based training (PBT) to optimize the hyperparameters of the PETS algorithm \citep{Chua2018PETS} during the training process.

In recent work, \citet{lai2021effective} theoretically analyzed the role of automated hyperparameter scheduling in Dyna-style MBRL, which reveals that a dynamic schedule of data ratio can be more effective than the fixed one. Motivated by such an analysis, they proposed the AutoMBPO framework to automatically schedule the key hyperparameters of the MBPO \citep{mbpo} algorithm. Empirical results show that MBRL methods with automatic hyperparameter optimization can achieve higher sample efficiency compared with those tuned manually \citep{zhang2021importance, lai2021effective}. How to better incorporate advanced techniques of AutoML \citep{hutter2019automated} into MBRL is a promising direction to investigate further.

\section{Applications of Model-based RL}

MBRL is of particular interest because of its potential to be applied in the real world, where a common feature is the intolerance of errors. This feature contradicts the fundamental mechanism, i.e., trial-and-error, of reinforcement learning methods. Therefore, between real-world applications and reinforcement learning, there must be a playground for training policies. The playground must have high fidelity to the real world and a high error tolerance for freely training reinforcement learning. 

Building hand-crafted simulators has been a widely adopted approach in cost-sensitive scenarios, such as autonomous driving \citep{zhou2021smarts}, industrial control \citep{hein2017benchmark}, decision optimization traffic control \citep{zhang2019cityflow}, electricity allocation \citep{vazquez2019citylearn} in smart cities, financial trading \citep{liu2020finrl}, control of tokamak plasma \citep{tokamak2022nature}, etc. A hand-crafted simulator can resemble the overall functionality of the real-world task, but it is hard to achieve high fidelity. The policy trained in a hand-crafted simulator may not be able to apply to the real-world task directly. This \emph{reality-gap} can be overcome by the meta-RL methods introduced above, where a meta-policy can be trained to adapt to the task on-the-fly. A single unrealistic simulator can also be used to facilitate policy training using some data collected in real-world task \citep{jiang2020nips}. Simulators are also useful for generating specific situations that are a rarity in real-world environments for learning robust policies \citep{zhou2021smarts,chou2018using,sun2021corner}.

Hand-crafted simulators are still expensive to build, costing huge expert time. Learning environment models from data can be a more efficient and lower-cost alternative to hand-crafted simulators. \cite{shi2019taobao} were the first to learn an environment model, Virtual-Taobao, for recommender systems. The environment model consists of detailed customer behaviors, i.e., click, buy, turn next page, and leave after reading the recommended items, which is believed to be extremely difficult to learn. It was shown in \citep{shi2019taobao} that the customer behavior can be well modeled by the proposed adversarial model learning method MAIL, which was later proved to have smaller compounding error as introduced in Sec. \ref{sec:model-learning-gail}. \cite{shang2021mlj} extended the MAIL method to model hidden factors, leading to a better model learning ability. Both the two studies show that the policies trained in the learned models can be deployed in the real-world tasks while maintaining similar performance as in the learned models, validated by real-world A/B tests.

From the above real-world applications, we also observe practical advantages of model-based methods:
\begin{itemize}
	\item \textbf{Full release of reinforcement learning power.} A simulator or a learned model allows any reinforcement learning algorithm to use sufficient explorations to train a good policy. Even if the model can be unrealistic, it is possible to constrain the exploration (e.g., \citep{shi2019taobao,tokamak2022nature}) to maintain the effectiveness of the learned policies.
	\item \textbf{Pre-deployment assessment.} It is crucial that a policy has been fully assessed before it can be deployed. However, assessing an improved policy is extremely difficult. Unlike supervised learning scenarios that commonly employ an identically distributed dataset to validate prediction models, an improved policy can easily derive a state-action distribution different from that of the collected data. The off-policy evaluation aims to evaluate a policy using historical non-identical distributed datasets. However, current off-policy evaluation methods have not shown reasonable effectiveness in benchmarks \citep{qin2021neorl}. While a recent study starts to combine off-policy evaluation and model-based policy evaluation \citep{jin2022arxiv}, running a policy in a simulator/model might be the most straightforward way to assess the performance.
	\item \textbf{Decision explanation.} Running a policy in a simulator/model can not only assess the performance of the policy but also allow us to see the specific decisions at states. These decisions themselves are also very useful for the decision-maker to evaluate the confidence of the policy in subjective ways. When the decisions show good rationality or even better ideas, the policy can gain more trust from the decision-maker, which is important in practice.
\end{itemize}

\section{Conclusions and Future Directions}
In this survey, we took a review on model-based reinforcement learning, which was a classic tabular RL method in the 1990s and has just got a renaissance for deep RL in recent years. Sample efficiency has always been the target to optimize in RL, particularly in the deep learning era. Model-based methods play an important role in achieving state-of-the-art sample efficiency in deep RL. From our survey, we have noticed some occurring developments of MBRL. We summarize the directions below:
\begin{itemize}
	\item \textbf{Learning generalizable models.} As a playground, models should be tolerant of the running of arbitrary policy. The generalization ability is the key to the success of MBRL. Recent progress includes introducing causal learning into model learning. A correct causal structure, \citep{zhu2022arxiv} as well as better modeling causal effect, \citep{chen2022arxiv} can help build good models. Causal model learning shows a way toward strong model generalization. 
	\item \textbf{Learning abstract models.} State abstraction \citep{Dietterich99} and temporal abstraction \citep{SuttonPS99} can map the original MDP to a low dimensional and compact MDP where the reinforcement learning task can be much simplified. Leveraging state and temporal abstraction, model-learning can happen in low dimensional space and thus becomes an easy task. We have observed some possibilities of learning abstract models \citep{Jiang2015ICML,zhu2022aaai}, while much more need to be done. Abstractions naturally lead to hierarchical reinforcement learning, which we find is almost an untouched topic.
	\item \textbf{Training generalizable policies.} Meta reinforcement learning, as discussed above, relies on model randomization and produces a meta-policy that can generalize to similar environments. The generalization ability of the meta-policy is rooted in model variations. However, the question of how to generate models such that the trained meta-policy adapts to the target environment is largely overlooked.
    \item \textbf{Model-based multi-agent RL.} Planning in various multi-agent scenarios with the multi-agent environment model is just in its infancy. By observing the recently emerging works, it is of high potential to incorporate model-based methods to improve the coordination among the team agents and increase the sample efficiency of training. More investigation shall be dedicated to multi-agent environment learning, planning, and communication mechanism design with the model.
    \item \textbf{Foundation models.} In the general machine learning community, learning foundation models is a recent emerging learning paradigm \citep{foundationmodels} that shows strong performance in a large range of vision and natural language processing tasks. This paradigm is also shifting in decision-making tasks by learning a single policy model \citep{generalistagent}. Other than foundation policy models, foundation environment models is a large region to be explored. 
\end{itemize}
Other possible directions include improving value discrepancy bounds, automatic scheduling in MBRL, adaptive model usage, life-long model learning \citep{Wu2020jaamas}, etc. It is reasonable to expect that there will be a series of breakthroughs towards more efficient and applicable RL techniques along the direction of model-based RL in the near future. 

\section*{Acknowledgement}
This work is supported by National Key Research and Development Program of China (2020AAA0107200) and National Natural Science Foundation of China (61876077,62076161).

\bibliographystyle{abbrvnat}
\bibliography{reference}

\end{document}